%% file: main.tex
\documentclass[10pt,twocolumn,letterpaper]{article}

\usepackage[pagenumbers]{cvpr} 

\usepackage{microtype}
\usepackage{graphicx}
\usepackage{multirow}
\usepackage{booktabs}
\usepackage{colortbl}
\usepackage{subfloat}
\usepackage{floatrow}
\usepackage{float}
\usepackage{pifont}

\newfloatcommand{capbtabbox}{table}[][\FBwidth]
\newfloatcommand{capbfigbox}{figure}[][\FBwidth]

\usepackage{algorithm}
\usepackage{algorithmicx}
\usepackage[noend]{algpseudocode}

\usepackage{pifont}
\usepackage{bbding}
\usepackage{caption}

\input{preamble}
\definecolor{cvprblue}{rgb}{0.21,0.49,0.74}
\usepackage[pagebackref,breaklinks,colorlinks,allcolors=cvprblue]{hyperref}

\renewcommand{\thefootnote}{\fnsymbol{footnote}}
\def\paperID{*****} 
\def\confName{CVPR}
\def\confYear{2026}

\title{GTR-Turbo: Merged Checkpoint is Secretly a Free Teacher\\for Agentic VLM Training}

\author{
Tong Wei$^{1}$\thanks{Work done during an internship at Tencent. Code can be found \href{https://github.com/weit123/GTR-Turbo}{here}.}
\quad Yijun Yang$^{2\ast}$
\quad Changhao Zhang$^{1}$
\quad Junliang Xing$^{1}$\footnotemark[3] \\
\quad Yuanchun Shi$^{1}$
\quad Zongqing Lu$^{3}$
\quad Deheng Ye$^{2}$\thanks{Corresponding Authors. $^{\ast}$Project Lead.} \\[0.5mm]
$^{1}$Tsinghua University \quad $^{2}$Tencent Hunyuan \quad $^{3}$Peking University\\
\tt\small \{wt22,zhangcha25\}@mails.tsinghua.edu.cn, yijun.steven.yang@gmail.com, \\ \tt\small \{jlxing,shiyc\}@tsinghua.edu.cn, zongqing.lu@pku.edu.cn, dericye@tencent.com
}

\begin{document}
\maketitle

\setcounter{footnote}{0}
\renewcommand{\thefootnote}{\arabic{footnote}}

\input{sec/0_abstract}    
\input{sec/1_intro}
\input{sec/2_related_works}
\input{sec/3_gtr_turbo}
\input{sec/4_exps}
\input{sec/5_conclusion}
\begin{flushleft}
\section*{Acknowledgement} 
This work is supported by the National Key Research and Development Plan under Grant No. 2024YFB4505500 \& 2024YFB4505503.
\end{flushleft}
{
    \small
    \bibliographystyle{ieeenat_fullname}
    \bibliography{main}
}

\input{sec/X_suppl}

\end{document}

%% file: sec/0_abstract.tex
\begin{abstract}
Multi-turn reinforcement learning (RL) for multi-modal agents built upon vision–language models (VLMs) is hampered by sparse rewards and long-horizon credit assignment.
Recent methods densify the reward by querying a teacher that provides step-level feedback, e.g., Guided Thought Reinforcement (GTR) \cite{wei2025gtr} and On-Policy Distillation \cite{lu2025onpolicydistillation}, but rely on costly, often privileged models as the teacher, limiting practicality and reproducibility.
We introduce \textbf{GTR-Turbo}, a highly efficient upgrade to GTR that matches its performance without training on or querying an expensive teacher model. 
Specifically, GTR-Turbo merges the weights of checkpoints produced during ongoing RL training and then uses the resulting merged model as a ``\textbf{free}'' teacher to guide subsequent RL via supervised fine-tuning or soft logit distillation. This design removes dependence on privileged VLMs (e.g., GPT or Gemini), mitigates the ``entropy collapse'' observed in prior work, and maintains stable training.
Across diverse visual agentic tasks, GTR-Turbo improves the accuracy of the baseline model by 10–30\% while reducing wall-clock training time by 50\% and compute cost by 60\% relative to GTR.\looseness-1

\end{abstract}

%% file: sec/1_intro.tex
\section{Introduction}
\label{sec:intro}

Vision-language models (VLMs) have evolved beyond simple static multi-modal question-answering systems, demonstrating the capability to perceive, reason, and act autonomously in interactive environments to achieve specific goals. Reinforcement learning with verifiable outcome rewards (RLVR) \cite{shao2024deepseekmath} enables such models to be fine-tuned directly through verifiable reward signals provided by the environment dynamics, effectively replacing learned reward models. This approach has shown remarkable success in domains such as mathematics and code generation \cite{openaio3, openaigpt5, guo2025deepseek, qwen3, gemeni2.5pro}. However, when applied to \textit{multi-turn agentic tasks}, vanilla RL methods often struggle due to sparse rewards, long-horizon trajectories, and noisy environments. These challenges can lead to incomplete, inconsistent, and low-diversity responses and actions, ultimately degrading performance, referred to as \textit{thought collapse}~\cite{wei2025gtr} or similar concepts (e.g., ``entropy collapse'') in many recent literatures \cite{shumailov2024ai, wang2025ragen, cui2025entropy}.

To this end, methods such as Guided Thought Reinforcement (GTR) \cite{wei2025gtr} and On-Policy Distillation \cite{lu2025onpolicydistillation} have introduced a new paradigm for multi-turn agent training by distilling knowledge from a larger and stronger teacher model that provides step- and even token-level guidance/supervision. They effectively regulate the agent's reasoning process and improve the quality of resulting actions. However, to achieve the most appealing performance, such a paradigm typically relies on expensive, often privileged models as the teacher, imposing severe scalability constraints, including high computational cost, longer training times, and the potential inaccessibility of cutting-edge models.

In this paper, we propose \textbf{GTR-Turbo}, a highly efficient solution to the above challenge. We highlight that \textbf{merging historical checkpoints generated throughout the RL training secretly creates a capable teacher for guidance}, completely free of additional training or external model dependency (see Figure~\ref{fig:framework} for a visualized explanation). This design not only preserves the automated reward mechanism, flexibility, and final performance of the original GTR method but also significantly accelerates training, reduces computational overhead, and thus achieves superior scalability.

Specifically, after each RL update \cite{schulman2017proximal}, we save the model weights and include them in our checkpoint buffer. By employing the TIES merging technique \cite{yadav2023ties}, we effectively avoid parameter interference among these models. The resulting merged model aggregates prior experience and consistently outperforms the current training agent, thereby serving as a teacher (see Figure~\ref{fig:model_merging_demo} for a proof-of-concept result). The guidance information can be used either through SFT-based online imitation learning, as in the original GTR framework, or via soft logit distillation with KL regularization, which replaces the teacher model's autoregressive generation with a single forward pass to further improve training efficiency and encourage exploration. As illustrated in Figure~\ref{fig:framework}, GTR-Turbo is a flexible, scalable, and self-evolving agent training method that does not require external supervision signals from the API model or human annotations, yet enables reliable and controllable decision-making in complex and challenging visual interactive environments.

Empirically, we post-train the Qwen2.5-VL-7B \cite{bai2025qwen2.5} and Qwen3-VL-8B \cite{Qwen3-VL} using GTR-Turbo. In the complex Points24 card game task \cite{zhai2025fine}, the resulting agent achieves state-of-the-art (SOTA) results, surpassing both GTR and other strong baselines while cutting 50\% training time, 100\% API calling, and 60\% of the compute cost, demonstrating remarkable efficiency improvement. In the widely adopted and more challenging ALFWorld visual environments \cite{shridhar2020alfworld}, where the observation only consists of images, an episode can span over 50 steps, and rewards are extremely sparse, typically provided at the end of tasks, GTR-turbo still makes rapid and stable progress on task success rate, outperforming many baselines with a comparable model size. Furthermore, we conduct comprehensive ablation studies to evaluate different design choices and configurations.\looseness-1

%% file: sec/2_related_works.tex
\section{Related Works}
\label{sec:related works}
\subsection{Agent Training for LLMs and VLMs}
Training agents that utilize general-purpose foundation models to solve specific decision-making problems has long been a central research topic. Early work primarily focuses on training-free prompting techniques or adding adapter modules to fixed base models \cite{wei2022chain, yao2023tree, yao2023react, wang2023describe, wang2023voyager, huang2022language, park2023generative, ahn2022can, shinn2023reflexion, wu2023autogen,zhou2024wall}. For multimodal tasks built upon VLM backbones, common approaches involve translating observations into textual descriptions or aligning their embeddings with LLMs \cite{ahn2022can, driess2023palm, gao2024physically, huang2023voxposer, mu2023embodiedgpt, sumers2023distilling, yang2024octopus, brohan2023rt,yang2024embodied}. However, such limited model adjustments struggle to cope with dynamic and complex environments, restricting the agent's robustness and adaptability.\looseness-1

More recent studies have focused on using RL to obtain improved agent policies through interaction with the environment. Beyond traditional PPO \cite{schulman2017proximal}, variants such as GRPO \cite{shao2024deepseekmath} and DAPO \cite{yu2025dapo} improve training stability and sample efficiency, while enabling long Chain-of-Thought reasoning and inference-time scaling, achieving strong performance on single-turn tasks such as math and code.

For multi-turn agentic tasks, concurrently with our work, a variety of large-scale RL training systems have emerged \cite{fu2025areal, wang2025ragen, zhang2025agentrl, li2025efficient, yu2026proact}. In the context of general visual reasoning, RL4VLM \cite{zhai2025fine} introduced a foundational framework that directly applies PPO to VLM post-training, providing a reference point for many subsequent studies \cite{wei2025gtr, wang2025vagen}.

\subsection{Process Guidance Providing Dense Rewards}
Sparse rewards have long been a core challenge in reinforcement learning. Prior research on deep-thinking LLMs has affirmed the critical role of process supervision in enhancing the logical consistency of reasoning \cite{shao2024deepseekmath}. One approach trains a Process Reward Model (PRM) to assess the reasoning process \cite{lightman2023let, uesato2022solving}, but requires costly human annotations to obtain high-quality data. Another solution focuses on credit assignment \cite{wang2024q, cui2025process, yuan2024free, feng2025group}, which decomposes the final reward into finer-grained signals to better attribute contributions across intermediate reasoning steps. Additionally, many studies have sought to enhance process guidance by leveraging large models, such as LLM-as-a-judge mechanisms \cite{zhang2024generative, gao2024llm, xia2024evaluating}, automated label generation \cite{wei2025gtr}, or world models to provide future information \cite{wang2025vagen}.

\subsection{Model Merging Techniques}
Merging the weights of multiple models is a well-established technique in machine learning for enhancing model performance \cite{yang2024model}. Studies have demonstrated that merging models trained on different downstream tasks can produce a unified and more versatile model \cite{ilharcoediting, yangadamerging, yu2024language}; combining models under varying hyperparameter settings can further boost performance \cite{wortsman2022model}; and merging historical checkpoints can help escape local optima while mitigating catastrophic forgetting \cite{huang2017snapshot, li2025temporal}. Recently, model-merging techniques have also seen rapid adoption in large language models, proving effective in both pre-training \cite{sanyalearly,li2025model} and post-training \cite{ilharcoediting,yu2024language,li2025temporal} stages.
Beyond simple averaging, studies have explored a range of additional model-merging techniques. Fisher Merging \cite{matena2022merging} estimates the importance of each parameter using the Fisher Information Matrix; Task Arithmetic \cite{ilharcoediting} constructs task vectors and performs arithmetic operations; TIES-merging \cite{yadav2023ties} introduces trimming and sign elections to mitigate parameter interference; and DARE \cite{yu2024language} incorporates random dropouts to enhance the effectiveness of multi-task model integration.\looseness-1

%% file: sec/3_gtr_turbo.tex
\section{The GTR-Turbo Framework}
\label{sec:gtr_turbo}
\subsection{Revisiting Guided Thought Reinforcement}
\begin{figure}[t]
  \centering
  \vspace{-5pt}
  \includegraphics[width=\linewidth]{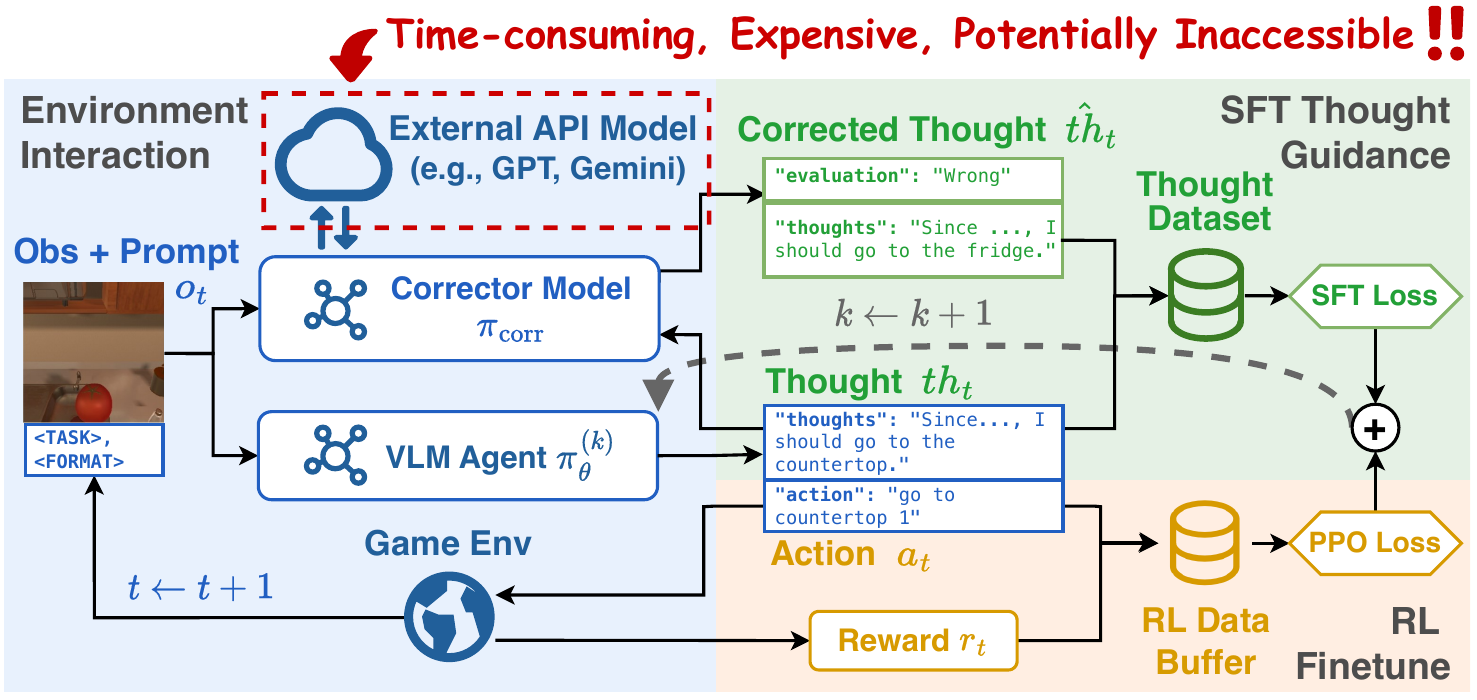}
  \caption{\textbf{Illustration of the original GTR framework \cite{wei2025gtr}.} It uses a multi-modal API model as the corrector, such as GPT or Gemini, to evaluate and refine the agent’s reasoning content (i.e., thought $th_{t}$) at each RL step, which is costly, time-consuming, and potentially inaccessible, constraining its own scalability.}
  \label{fig:old_framework}
\end{figure}

Guided Thought Reinforcement (GTR) \cite{wei2025gtr} is a flexible reinforcement learning framework designed for training VLM agents in multi-turn decision-making tasks. As shown in Figure~\ref{fig:old_framework}, it leverages a VLM-as-a-corrector mechanism, where an external VLM model acts as a corrector to evaluate and refine the agent's reasoning at each RL step. By jointly performing SFT on reasoning tokens and RL on action tokens simultaneously, GTR effectively solves ``thought collapse'' caused by the absence of reliable thinking rewards. GTR also incorporates Dataset Aggregation (DAgger) \cite{ross2011reduction} to mitigate the distribution shift issue that arises during the dynamic RL training. \looseness-1

Formally, we denote the agent's observation as $o$, thought output as $th$, and action as $a$, given agent model $\pi_\theta$ and corrector model $\pi_{\text{corr}}$. $\mathcal{B}$ represents the PPO data buffer and $\mathcal{D}$ denotes the thought data buffer. If we term $[l]$ as the $l$-th token and $[<l]$ as the first $l$ tokens, then the objective of GTR can be represented as:
\vspace{-2pt}
\begin{equation}
\label{eqn:GTR_objective}
    \min_{\theta} \mathop{\mathbb{E}}_{(o,a)\sim\mathcal{B}}\!\!\!\mathcal{L}_{\text{PPO}}(o, a)+\mathop{\mathbb{E}}_{(o,th)\sim\mathcal{D}}\!\!\!\mathcal{L}_{\text{SFT}}(o, \pi_{\text{corr}}(o, th)),
\end{equation}
\vspace{-2pt}
where
\vspace{-2pt}
\begin{align}
\label{eqn:GTR_loss}
\mathcal{L}_{\text{PPO}}(o, a) &= - \min \left(\frac{\pi_\theta(a|o)}{\pi_{\theta_\text{old}}(a|o)} A^{\pi_{\theta}}(o,a), \right. \nonumber \\ 
& \left.\text{clip}\left({\frac{\pi_\theta(a|o)}{\pi_{\theta_\text{old}}(a|o)}, 1-c, 1+c}\right) A^{\pi_{\theta}}(o,a)\right), \nonumber \\
\mathcal{L}_{\text{SFT}}(o, th) &= -\sum_l \log \pi_\theta(th_{[l]}|o, th_{[<l]}).
\end{align}
\vspace{-5pt}

Although GTR has achieved remarkable progress across multiple tasks, it comes with significant prerequisites and costs. First, GTR requires a larger and more powerful external model to ensure correctness and serve as a reliable teacher. However, such models for downstream domains are sometimes inaccessible in practice, and their capabilities directly affect training quality. Moreover, when using closed-source models such as GPT or Gemini as teachers, the need for step-level online API calls significantly slows training and incurs substantial costs. As shown in Table \ref{tab:GTR_overhead}, using GPT-4o (even a lightweight model in the GPT family) as the teacher requires approximately 4 days and costs approximately 150 USD to post-train LLaVA-1.6-7B with GTR for 15,000 steps. Using smaller teacher models can reduce overhead but degrade final performance, even failing to provide meaningful guidance.

\begin{table}[htb]
    \centering
    \resizebox{0.9\linewidth}{!}{
    \begin{tabular}{cccc}\toprule
    \textbf{Corrector Model} & \textbf{Performance} & \textbf{Token Usage (Cost)}  & \textbf{Time}  \\ \midrule
    GPT-4o & 17.5\% & 33.5M ($\sim$\$146.56) & 86h \\
    Qwen2.5-VL-72B & 6.5\% & 33.8M ($\sim$\$18.59) & 110h \\
    Qwen2.5-VL-7B & 0\%* & 31.2M ($\sim$\$4.29) & 56h \\ \bottomrule
    \end{tabular}
    }
    \caption{\textbf{Training time and token usage of the GTR framework.} Experiments to train the LLaVA-v1.6-mistral-7B model for 15,000 steps on the Points24 task, using different models as the corrector. * - The corrector model fails to provide valid thought guidance.}
  \label{tab:GTR_overhead}
\end{table}

\begin{figure*}[t]
  \centering
  \includegraphics[width=0.95\linewidth]{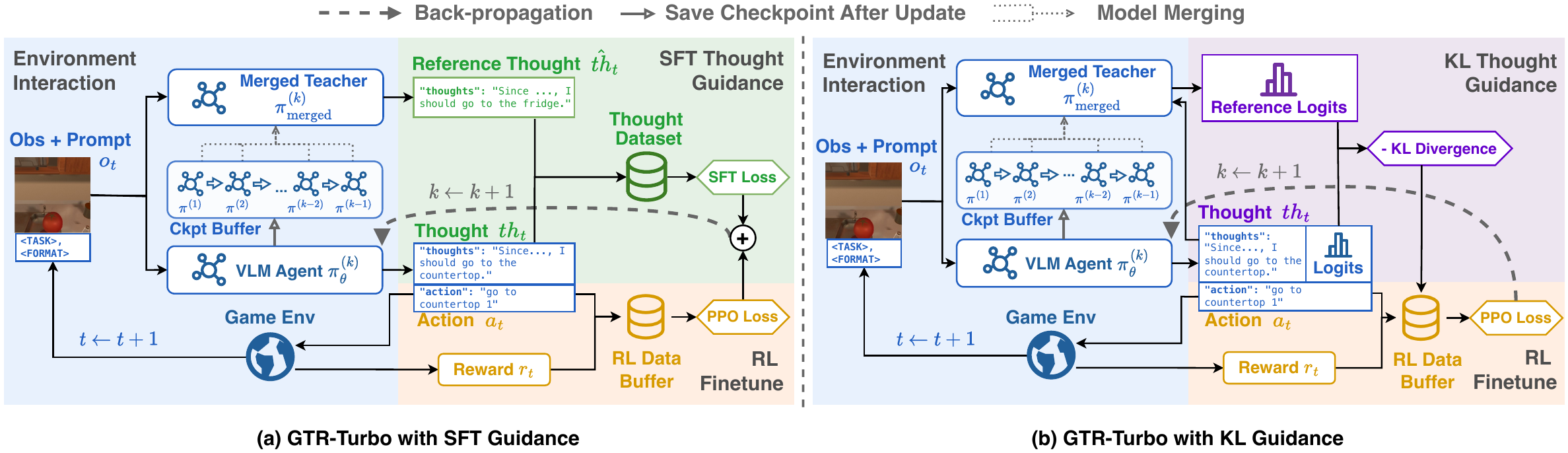}
  \caption{\textbf{Overview of the GTR-Turbo framework.} Beyond the GTR training of VLM agents (Figure \ref{fig:old_framework}), GTR-Turbo stores historical checkpoints and merges them into a teacher model (blue region), and then incorporates the PPO update (orange region) with thought guidance by minimizing either SFT loss (green region) or KL divergence (purple region), enabling flexible, scalable, and self-guided agentic RL training.\looseness-1}
  \label{fig:framework}
  \vspace{-8pt}
\end{figure*}

In this paper, we introduce an elegant solution to the above limitation: \textbf{merging the historical checkpoints generated during the RL training constructs a teacher model for free}, as shown in Figure \ref{fig:framework}, thereby eliminating the dependence on expensive external models. Empirical results demonstrate that this approach can achieve comparable and even superior performance while dramatically reducing both training time and token cost.\looseness-1

\subsection{Merged Checkpoints as the Teacher}
\label{sec:model_merging}
As an efficient model adaptation method, model merging has been widely adopted in the post-training stage. It includes merging heterogeneous models trained on different downstream tasks, enabling continual learning and capability expansion \cite{ilharcoediting, yangadamerging, yu2024language}, or merging homogeneous models trained on the same task, achieving stronger overall performance \cite{wortsman2022model, huang2017snapshot, li2025temporal}. In GTR-Turbo, we design a buffer that comprises historical model checkpoints as the agent's RL training progresses. The merged model in the $k$-th update epoch is formulated by Equation \ref{eqn:model_merging}:
\vspace{-2pt}
\begin{equation}
\label{eqn:model_merging}
    \pi^{(k)}_\text{merged} = \sum_{i=1}^{k-1} w_i\pi_{\theta}^{(i)}.
\end{equation}
\vspace{-2pt}

\begin{figure}[htbp]
    \vspace{-0.5em}
    \centering
    \includegraphics[width=0.9\linewidth]{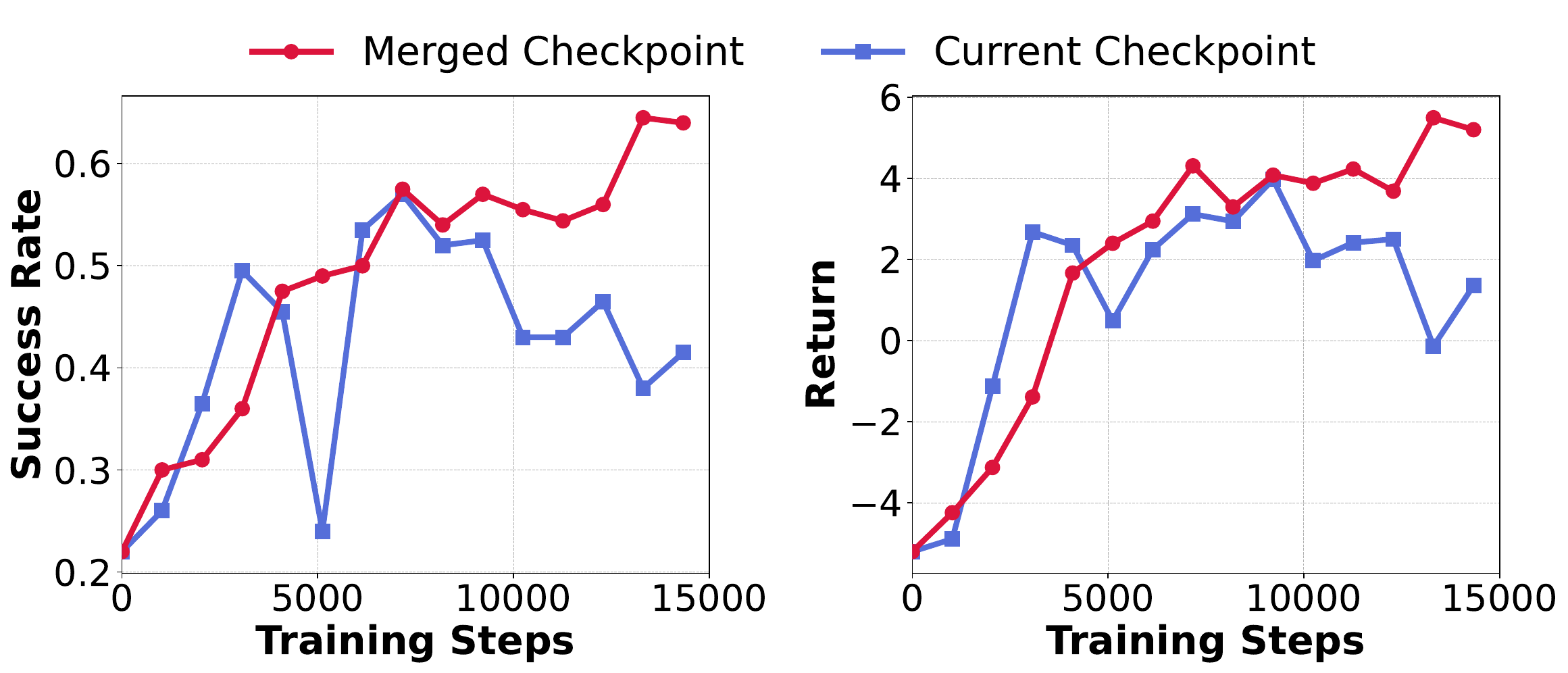}
    \caption{\textbf{The performance comparison of the merged checkpoint and the current checkpoint on Points24.} We adopt the Qwen2.5-VL-7B as the base model and highlight that model merging leads to a stronger and more stable agent $\pi^{(k)}_{\text{merged}}$ (red line) that can serve as a teacher to guide the following RL for training $\pi_{\theta}^{(k)}$.}
    \vspace{-5pt}
    \label{fig:model_merging_demo}  
    \vspace{-3pt}
\end{figure}

The merged teacher requires no additional training and yields a better model by optimizing over a smoother loss surface while effectively preserving past experiences. To validate this statement, we use a checkpoint trajectory produced by training Qwen2.5-VL-7B on Points24 with GTR. At each update, we evaluate both the current model and the merged model obtained from all preceding checkpoints. As shown in Figure \ref{fig:model_merging_demo}, the merged model is more stable and achieves better performance, serving as a capable teacher.

\paragraph{Merging Method} Directly merging all parameters of checkpoints can introduce harmful interference, where changes in redundant parameters affect the model's performance after merging. To avoid this issue, we adopt the \textit{Trim, Elect Sign, and Merge (TIES)} method \cite{yadav2023ties}, which consists of three steps, (1) Trimming: redundant parameter changes are removed by retaining only those with magnitudes in the top-$k$\%; (2) Sign election: for each parameter, we compute the total magnitude of its positive and negative values across all models and apply a majority vote to determine the elected sign vector; (3) Selective averaging: only parameters whose signs match the elected sign are included in the merging computation. This procedure mitigates the influence of minor perturbations, ensuring a tractable merging process.

\paragraph{Weight Adjustment Variants} Adjusting weights for every checkpoint results in different variants of merging methods. We study two commonly used strategies in this work: Simple Moving Average (SMA) and Exponential Moving Average (EMA). SMA treats all checkpoints equally and computes the arithmetic mean of them. EMA prioritizes more recent checkpoints by applying a sequence of decayed weights with a smoothing factor $\alpha$:

\vspace{-1em}
\begin{align}
\label{eqn:weight_variants}
    \pi_\text{merged}^{(k)} &= \frac{1}{k-1}\sum_{i=1}^{k-1} \pi_{\theta}^{(i)}, &\text{(SMA)} \nonumber \\
    \pi_\text{merged}^{(k)} &= \alpha \cdot \pi_{\theta}^{(k-1)} + (1-\alpha) \cdot \pi_\text{merged}^{(k-1)} .&\text{(EMA)}
\end{align}

\subsection{Thought Guidance via Supervised Fine-tuning}
After merging checkpoints, we can replace the corrector model (red dashed box in Figure \ref{fig:old_framework}) in the original GTR framework with the new merged teacher model. Since the teacher and agent are homologous, they take the same input. Similar to GTR, GTR-Turbo (SFT) implements the guidance by minimizing the SFT loss between thought tokens generated by two models, as demonstrated in Figure \ref{fig:framework} (a).\looseness-1

At each RL step, after the agent generates its action and thought given the observation, the same context is provided to the teacher to produce a reference thought, which is then stored in a replay buffer. In subsequent PPO updates, we sample data pairs from the thought to compute the SFT loss, which is added to the original PPO loss for backpropagation. Similar to GTR, we also incorporate format rewards and DAgger \cite{ross2011reduction} techniques to further stabilize training. If $\hat{th}$ represents the thought output of the teacher model $\pi_\text{merged}(o)$, the optimization target can be written as:\looseness-1
\begin{equation}
\label{eqn:sft_objective}
    \min_{\theta} \mathop{\mathbb{E}}_{(o,a)\sim\mathcal{B}}\!\!\!\mathcal{L}_{\text{PPO}}(o, a)+\mathop{\mathbb{E}}_{(o,\hat{th})\sim\mathcal{D}}\!\!\!\mathcal{L}_{\text{SFT}}(o, \hat{th}).
\end{equation}

\subsection{Soft Logit Distillation via Minimizing Reverse KL Divergence} 
As discussed in the prior section, GTR-Turbo (SFT) adopts an SFT loss as the online imitation objective. However, this formulation requires autoregressive generation from the teacher, which incurs substantial computational overhead. To improve efficiency, we replace the SFT objective with a KL-based guidance that only requires parallel inference: we compute the negative KL divergence between the agent and the teacher as thought reward, encouraging the agent to align its token-level output distribution with that of the teacher. This soft logit distillation imposes a more relaxed constraint on the student, significantly accelerates training, and improves performance.

\textbf{Using KL divergence offers non-trivial advantages.} First, since it is grounded in the model's logit outputs, it is almost unhackable. A smaller KL value indicates closer alignment between the agent and teacher outputs, with zero achieved when they are identical. Second, KL divergence captures probability information over all candidate tokens, whereas an SFT label is one-hot supervision for the target token. Finally, computing the KL divergence requires only a single forward pass, making it highly efficient. This KL variant also eliminates the need for an additional thought dataset in GTR, thereby reducing memory consumption.

Previous research \cite{guminillm,wu2025rethinking,lu2025onpolicydistillation} has proven the advantages of using reverse KL for knowledge distillation. As shown in Figure \ref{fig:framework} (b) and Equation \ref{eqn:kl_objective}, given the thought output, we compute the reverse KL between the agent and the teacher, average over all tokens, and take its negative value as an auxiliary reward for PPO updates, which is a required approach for multi-step RL optimization. Since this sentence-level KL estimation may yield negative values and produce misleading reward signals, we clip the negative parts to ensure the reward's validity (see Section \ref{sec:ablation} for a detailed analysis).
\vspace{-3pt}
\begin{equation}
    \label{eqn:kl_objective}
    \max_{\theta} \mathop{\mathbb{E}}_{(o,(th,a))\sim\mathcal{B}} \left[\min\left(rA', \text{clip}\left(r, 1-c, 1+c\right) A'\right)\right],
\end{equation}
in which,
\begin{align}
    & r = \frac{\pi_\theta(a|o)}{\pi_{\theta_\text{old}}(a|o)}, \quad A' = A^{\pi_{\theta}}(o,a) - \text{RevKL}(\pi_\theta, \pi_\text{merged};th), \nonumber \\
    &\text{RevKL}(\pi_\theta, \pi_\text{merged};th) =  \mathbb{E}_l\left[\log\pi_\theta(th_{[l]}|th_{[<l]}) \right. \nonumber \\
    & \quad\quad\quad\quad\quad\quad\quad\quad\quad\quad\quad\quad \left.- \log\pi_\text{merged}(th_{[l]}|th_{[<l]})\right]. \nonumber
\end{align}

%% file: sec/4_exps.tex
\section{Experiments}
\label{sec:exps}

\begin{figure*}[t!]
\begin{floatrow}
\ffigbox[0.60\textwidth]{%
\centering
\includegraphics[width=0.9\linewidth]{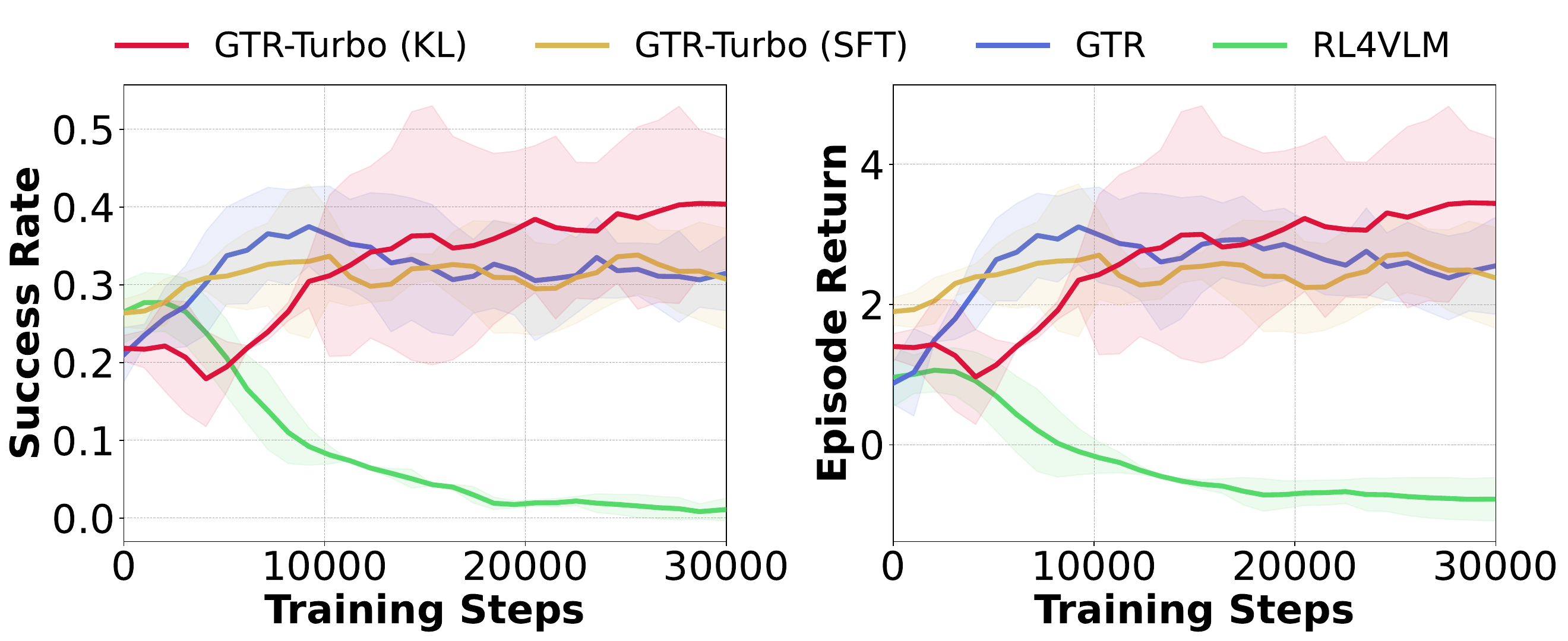}
}{%
  \caption{\textbf{Training curves on the Points24 game environment.} While GTR benefits from external knowledge in the early stage, our GTR-Turbo framework is also able to maintain a rational reasoning process and ultimately achieves the best overall performance. All curves are smoothed for better readability. All experiments employ the early-truncation strategy introduced by GTR for a fair comparison.}
  \label{fig:main_points24}
}
\hspace{-15pt}
\capbtabbox[0.38\textwidth]{%
    \centering
    \resizebox{0.64\linewidth}{!}{
        \begin{tabular}{ccc}\toprule
    \textbf{Model} & \textbf{SR(\%)}  & \textbf{ER}  \\ \midrule
    CNN+RL* & 0 & -1.12 \\
    GPT-4o & 2.5 & -6.35 \\
    GPT-4o + Tool & 13.5 & -3.59 \\
    Qwen2.5-VL-72B & 5.6 & -5.69 \\
    Qwen2.5-VL-32B & 0 & -7.25 \\
    Qwen2.5-VL-7B-sft & 22.0 & -3.2 \\
    RL4VLM & 3.5 & -13.3 \\
    GTR & 44.5 & 0.53 \\	
    GTR-Turbo (SFT) & 48.0 & 1.32 \\
    \textbf{GTR-Turbo (KL)} & \textbf{53.5} & \textbf{2.39}  \\ \bottomrule
    \end{tabular}
    }
}{%
  \caption{\textbf{Evaluation result of different models on the Points24 task.} GTR-Turbo significantly outperforms other RL training methods and commercial models in both success rate (SR) and episode return (ER). * - Reported in previous work.}
  \label{tab:evaluation_points24}
}
\end{floatrow}
\vspace{-10pt}
\end{figure*}

\subsection{Experimental Setup}
\paragraph{Environments} We conduct our experiments on two widely used and challenging visual agentic benchmarks: Points24 \cite{zhai2025fine} and ALFWorld \cite{shridhar2020alfworld}.

In Points24, the model must first perform fine-grained poker card recognition, followed by logical reasoning. At each step, the agent decides which number (1-10) or operator (e.g., $+,-,\times,\div$) to append to the current formula, ultimately forming the one equal to 24. Episodes sometimes involve more than 10 steps and require domain-specific skills, such as arithmetic computation, making them complex.
ALFWorld is a multimodal embodied simulator featuring diverse household tasks. The well-trained agent is expected to navigate in unfamiliar environments, locate and interact with objects to reach certain goals, which poses substantial challenges in visual perception, long-horizon planning, and commonsense reasoning. The length of ALFWorld tasks can exceed 50 steps, and each step may involve more than 20 possible actions, creating a huge action space that surpasses a large portion of existing agentic VLM benchmarks, including many GUI device control environments \cite{rawles2023androidinthewild, rawles2024androidworld, xie2024osworld}.
The reward signals in both environments are sparse. This makes them significantly more difficult than tasks with heuristic/oracle process rewards.

\paragraph{Baselines} We select other competitive methods for multi-turn VLM agent RL training as our baselines. RL4VLM \cite{zhai2025fine} directly applies PPO optimization on raw environment rewards. GTR \cite{wei2025gtr} introduces thought guidance via an external GPT-4o corrector model for knowledge distillation, enabling rapid improvement and representing the current SOTA method. In addition, we also compare the agent's final performance with several privileged API models.

\paragraph{Training Details} Compared to previous work, we adopt a stronger backbone model, Qwen2.5-VL-7B. We start from an SFT-initialized model that exhibits reasonable instruction-following capabilities, consistent with prior studies. The full configuration details are provided in the appendix.
To better evaluate training stability, we train the agent for 30,000 steps on Points24 and 20,000 steps on ALFWorld, which are 2x and 4x the budgets reported in previous work, respectively. We use two NVIDIA GPUs with 40GB of memory, one for the merged checkpoint teacher and the other for the LoRA \cite{hu2022lora} fine-tuned agent.

\begin{figure*}[!t]
\begin{floatrow}
\ffigbox[0.38\textwidth]{%
\centering
\includegraphics[width=0.9\linewidth]{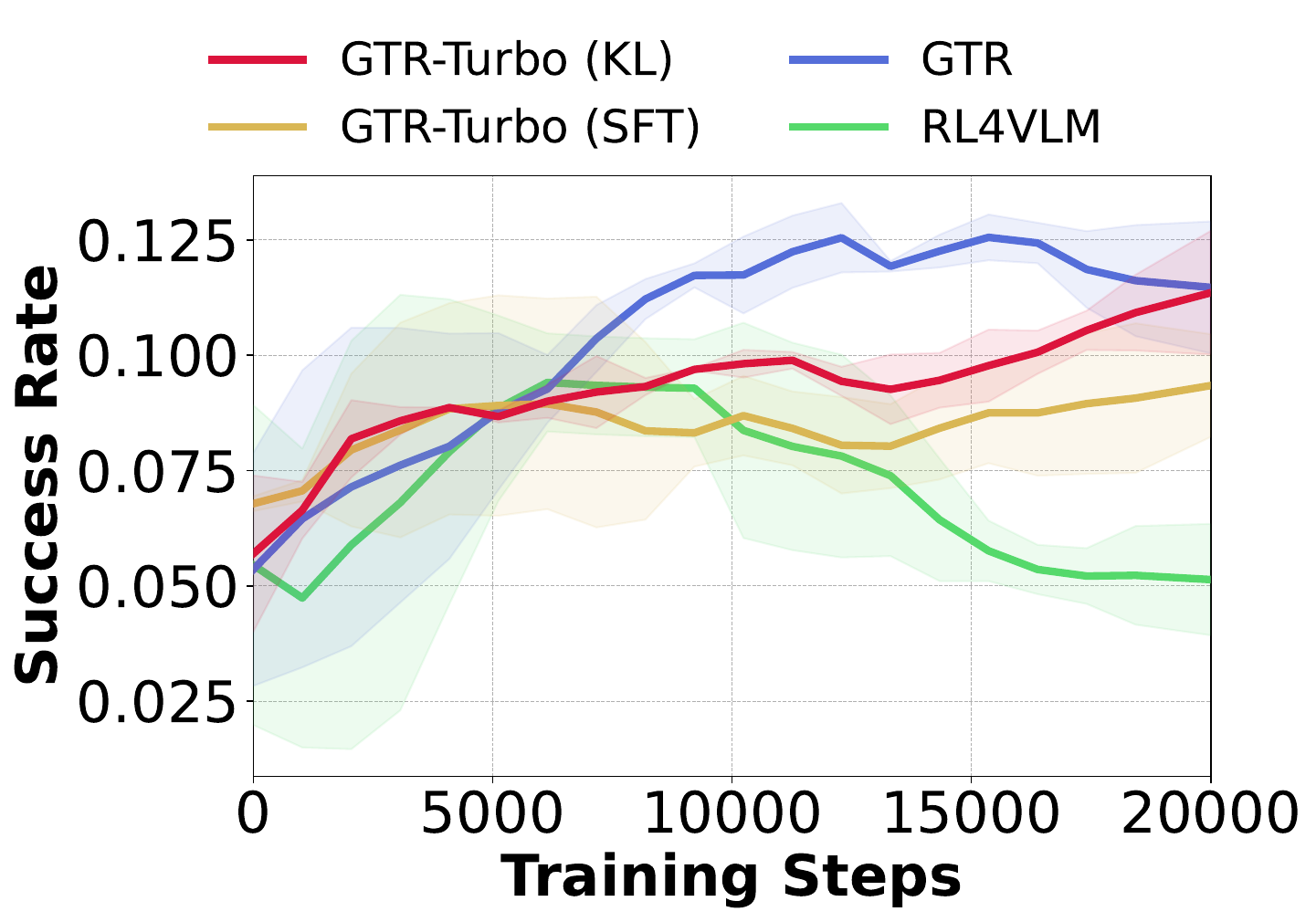}
}{%
  \caption{\textbf{Comparison of training curves in the ALFWorld environment.} Without relying on any powerful external models, GTR-Turbo achieves comparable performance purely through its own exploration, experience, and thought guidance.}
  \label{fig:main_alfworld}
}
\hspace{-15pt}
\capbtabbox[0.6\textwidth]{%
    \centering
    \resizebox{0.85\linewidth}{!}{
        \begin{tabular}{ccccccccc}\toprule
     \textbf{Model} & \textbf{Average} & \textbf{Pick} & \textbf{Clean} & \textbf{Heat} & \textbf{Cool} & \textbf{Look} & \textbf{Pick2} \\ \midrule
    CNN+RL* & 0 & 0 & 0 & 0 & 0 & 0 & 0 \\
    LLaMA-Adapter* & 0.13 & 0.17 & 0.10 & 0.27 & 0.22 & 0 & 0 \\
    GPT-4o & 0.42 & 0.53 & 0.67 & 0.22 & 0 & 0.25 & 0.47 \\
    Qwen-2.5-VL-72B & 0.32 & 0.38 & 0.14 & 0.50 & 0.27 & 0.63 & 0.21 \\
    Qwen-2.5-VL-32B & 0.21 & 0.44 & 0.25 & 0 & 0.2 & 0.16 & 0.2 \\
    Qwen-2.5-VL-7B-sft & 0.08 & 0.26 & 0.07 & 0.09 & 0 & 0.05 & 0 \\
    RL4VLM & 0.08 & 0.40 & 0 & 0 & 0 & 0 & 0 \\
    GTR & 0.16 & 0.44 & 0.14 & 0 & 0.14 & 0.05 & 0 \\ 
    GTR-Turbo (SFT) & 0.12 & 0.36 & 0 & 0.13 & 0 & 0.125 & 0 \\ 
    GTR-Turbo (KL) & 0.15 & 0.40 & 0.09 & 0 & 0.14 & 0.08 & 0.07 \\ \bottomrule
    \end{tabular}
    }
}{%
  \caption{\textbf{Comparison of success rates across different models in the ALFWorld environment.} We present the peak performance in the training curve for RL methods. GTR-Turbo achieves the same task success rate compared to GTR with significantly less training time and lower computational cost, maintaining excellent performance under its model scale. * - Reported in previous work.}
  \label{tab:evaluation_alfworld}
}
\end{floatrow}
\vspace{-10pt}
\end{figure*}

\subsection{Effectiveness of the GTR-Turbo Framework}
\paragraph{Points24}
Figure \ref{fig:main_points24} demonstrates the training curves of all methods. RL4VLM suffers from thought collapse, where the outputs become repetitive, incoherent, and templated. Consequently, the task success rate and episode return rapidly decline until the agent fails to succeed. By introducing an external GPT-4o model, GTR enables rapid knowledge distillation, leading to improvements in the early stages. However, as training progresses, the fixed external model cannot accumulate additional knowledge, thus limiting further learning.
In contrast, our GTR-Turbo, without any external knowledge, also achieves stable and consistent improvement purely through environmental feedback. Ultimately, the SFT-guided method reaches performance comparable to GTR, while the KL-guided version further surpasses all baselines.

In Table \ref{tab:evaluation_points24}, we present the final evaluation results of agents on the Points24 task, which are either trained using different RL methods or built upon privileged API models. GTR-Turbo achieves the best performance. The fine-tuned smaller model can easily outperform general-purpose models that are more than 10 times larger on tasks that require specialized domain knowledge. GTR-Turbo thus offers a promising approach to the customization/personalization of VLM agents.

\paragraph{ALFWorld} As illustrated in Figure \ref{fig:main_alfworld} and Table \ref{tab:evaluation_alfworld}, the baseline RL4VLM still exhibits model collapse, which undermines its training effectiveness. In such complex navigation tasks, advanced API models demonstrate substantially superior capabilities and thus can provide richer and more accurate guidance for GTR, enabling a rapid early increase. A closer analysis shows that external knowledge markedly reduces the need for exploration, allowing the agent to directly imitate correct trajectories.
While GTR-Turbo lacks access to such extensive knowledge and instead learns from its own exploration experience under the guidance of a merged teacher, the results show that the agent can learn independently and effectively. Even under this unfair comparison setting, GTR-Turbo (KL) obtains appealing scores on par with GTR while offering better efficiency. These findings highlight the potential of GTR-Turbo for training agents in hard tasks where off-the-shelf teachers may not exist.

\begin{table}[b]
\centering
    \resizebox{0.9\linewidth}{!}{
    \begin{tabular}{cccccc}\toprule
    \textbf{Env} & \textbf{Method} & \textbf{SR} & \textbf{Time} & \textbf{Cost Estimation} \\ \midrule
     \multirow{4}{*}{P24} & RL4VLM & 4\% & 86h & \$0 \\
      & GTR & 41\% & 191h  & \$307.78 / 70.35M tokens  \\ 
      & GTR-Turbo (SFT) & 48\% & 168h & \$216.72* & \\
      & \textbf{GTR-Turbo (KL)} & \textbf{54\%} & \textbf{89h} & \textbf{\$114.81}* & \\ \midrule
      \multirow{4}{*}{ALF} & RL4VLM & 8\% & 70h & \$0 \\
      & GTR & \textbf{16\%} & 164h & \$145.76 / 30.94M tokens  \\ 
      & GTR-Turbo (SFT) & 12\% & 118h & \$152.22* & \\
      & \textbf{GTR-Turbo (KL)} & 15\% & \textbf{78h} & \textbf{\$100.62}* & \\\bottomrule
    \end{tabular}
    }
    \caption{\textbf{Computation Time and Cost Comparison.} GTR-Turbo has comparable or even superior performance to GTR with significantly shorter training time and lower monetary cost. Reported costs account only for \textit{additional} overhead (excluding the base cost of agent training) and may fluctuate with market conditions. P24 - Points24, ALF - ALFWorld, SR - task success rate, * - Estimation based on the deployment cost of an additional GPU.}
    \label{tab:cost_comparison}
\end{table}

\begin{figure*}[t]
\begin{floatrow}
\ffigbox[0.32\textwidth]{%
\centering
\includegraphics[width=0.9\linewidth]{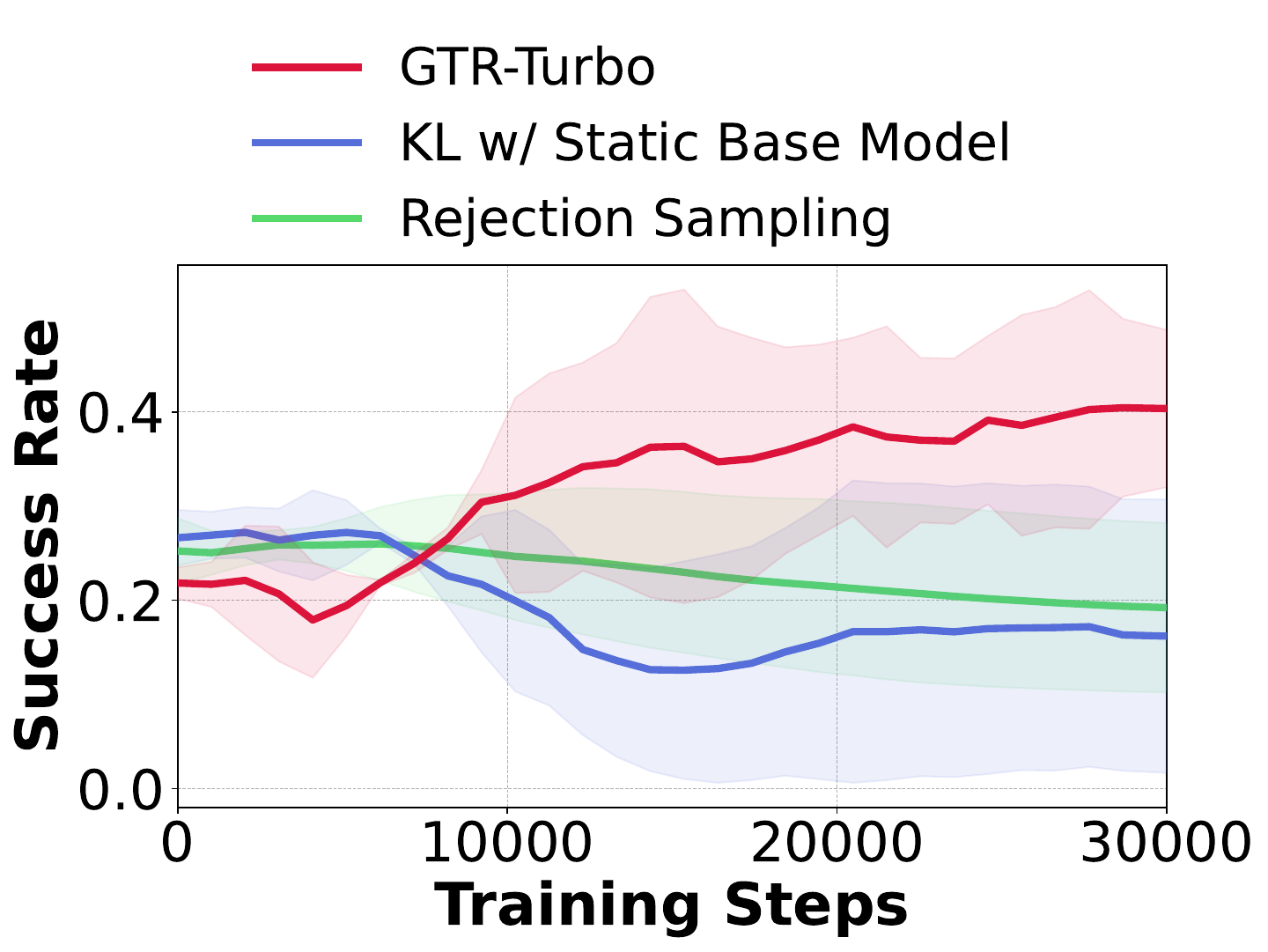}
}{%
  \caption{\textbf{Comparison with other self-improvement baselines.} The advantage over static-model KL regularization shows the necessity of model merging. The comparison with Rejection Sampling highlights the critical role of RL exploration.}
  \label{fig:ablation_method} 
}
\hspace{-15pt}
\ffigbox[0.32\textwidth]{%
\includegraphics[width=0.9\linewidth]{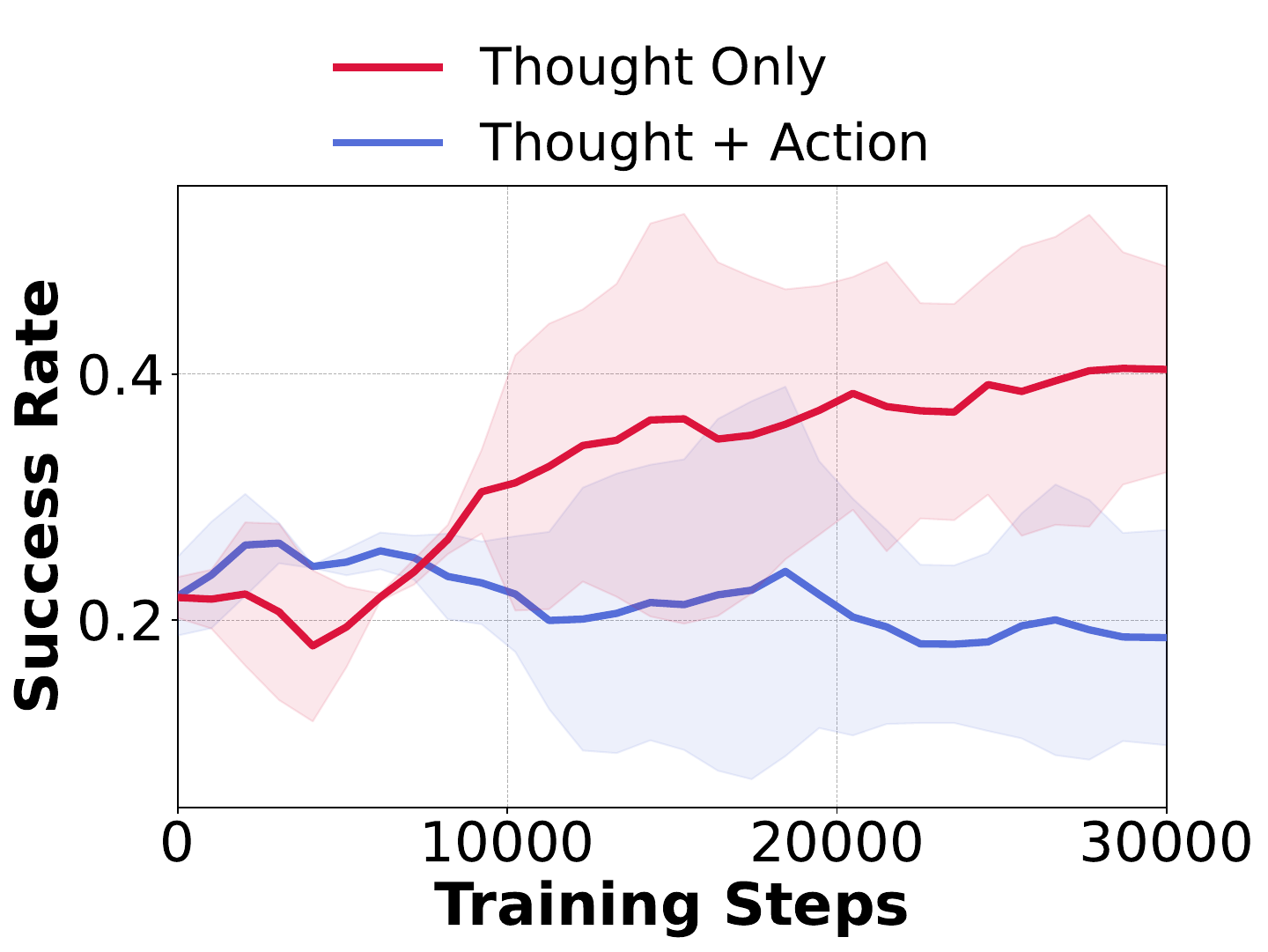}
}{%
  \caption{\textbf{Comparing different ranges of guidance.} Guiding full responses, including both the thoughts and actions simultaneously, is less effective, primarily because it limits the model's exploration, a process that is crucial for self-evolution in GTR-Turbo.}
  \label{fig:ablation_range} 
}
\hspace{10pt}
\ffigbox[0.32\textwidth]{%
\includegraphics[width=0.9\linewidth]{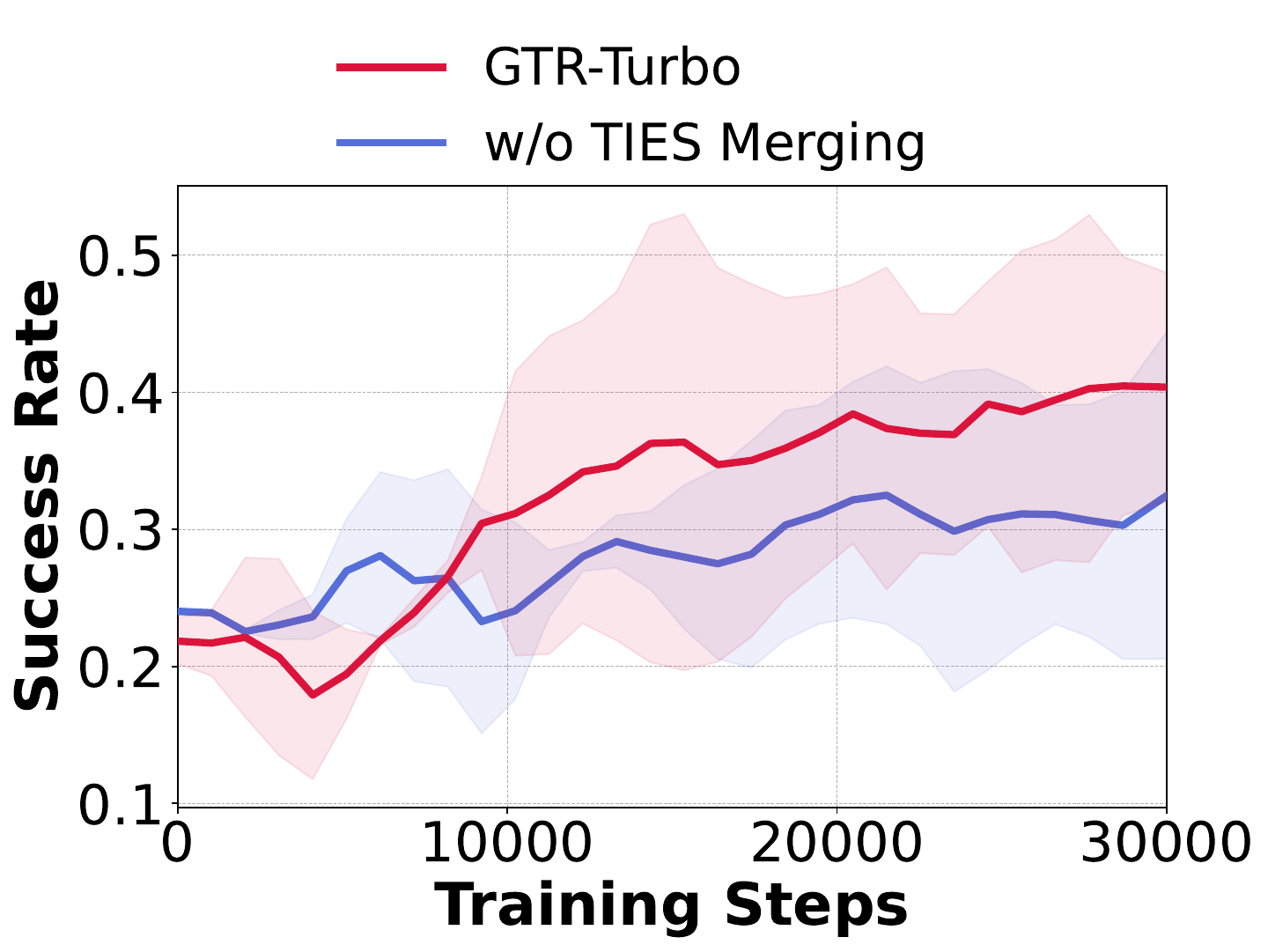}
}{%
  \caption{\textbf{Performance comparison with and without TIES merging.} The results demonstrate the robustness of TIES in the merging process, effectively enhancing the quality of the teacher model and improving the overall training gains.}
  \label{fig:ablation_ties} 
}
\end{floatrow}
\vspace{-15pt}
\end{figure*}

\paragraph{Computation Time and Cost} Table \ref{tab:cost_comparison} provides a comparison of the overhead across training methods. The cost estimates include only the \textit{additional} expenses required to implement each training mechanism and exclude the base cost of fine-tuning the VLM agent itself. For GTR, we compute the cost based on the token count via API calls. For GTR-Turbo, which requires an additional GPU to deploy the teacher model, we estimate the training cost by multiplying the hourly deployment cost per GPU by the total training time.
Note that both OpenAI API pricing and GPU costs can change over time. Moreover, different API providers may charge different prices due to their hardware configurations. which further corrupts the precision of our cost estimates.

We assume that the original RL4VLM framework incurs no additional overhead but exhibits poor training performance. GTR leverages large external models such as GPT-4o to guide RL training, but the resulting API costs are computationally prohibitive. Additionally, this approach introduces network latency and raises data privacy concerns. As a result, GTR requires much longer training time and also considerable expense, both of which limit its real-world applicability.

Our proposed GTR-Turbo is an elegant solution to this dilemma. By replacing costly external model calls with local inference, the SFT-guided GTR-Turbo already achieves noticeable reductions in overall time and cost. The KL-guided variant further reduces the total training time to that of RL4VLM, roughly half that of GTR. It also reduces the computational cost to as low as 40\% of that of GTR. Moreover, in scenarios where cutting-edge models are inaccessible or data privacy is critical, the fully self-contained and locally deployable GTR-Turbo offers an unparalleled advantage.

\subsection{Ablation Studies and Discussions}
\label{sec:ablation}

\paragraph{Comparison with other self-improvement baselines} We conducted a study to compare the effects of our model merging and the regularization trick commonly used by RLVR. As shown in Figure ~\ref{fig:ablation_method}, using the initial checkpoint (static base model) as the KL reference fails to achieve stable improvement as GTR-Turbo does.
We also compare GTR-Turbo with a simple yet provenly effective self-improvement baseline, Rejection Sampling \cite{yuan2023scaling}, which uses SFT on self-generated successful trajectories and thus does not rely on external models.
While simpler and faster, Rejection Sampling cannot solve the challenging Point24 as it struggles to generate the correct trajectories that the VLM agent can imitate. On the contrary, RL has the potential to explore better actions leveraging the reward feedback.

\paragraph{Range of Guidance} In Figure \ref{fig:ablation_range}, we try guiding the agent's full response, including both the thought and action. Consistent with observations from GTR, this approach is less effective. Combined with earlier results showing the advantage of KL-based guidance over SFT, we argue that for a self-contained system like GTR-Turbo, efficient exploration is critical as it directly determines the diversity of feedback and experience. Guiding actions or imposing stronger SFT constraints may improve the agent's ability to imitate the teacher, but at the cost of restricting the agent's exploratory freedom, limiting overall ability to adapt to the environment.

\paragraph{Effectiveness of TIES Merging} We experiment between TIES merging and the traditional linear averaging method \cite{ilharcoediting} to validate the effectiveness of this technique. Results in Figure \ref{fig:ablation_ties} show that TIES can indeed boost performance by mitigating the interference of redundant parameters and sign disagreement, allowing the merged model to better preserve and integrate learned capabilities. Meanwhile, linear merging also yields reasonable training gains, confirming the validity of the merged checkpoint as a teacher.

\begin{figure*}[t]
  \centering
  \includegraphics[width=0.9\linewidth]{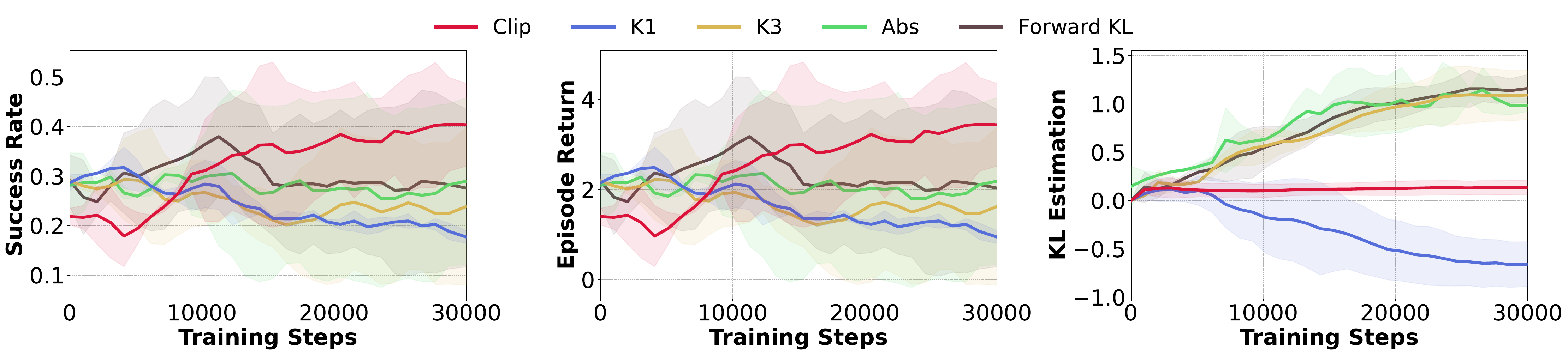}
  \caption{\textbf{Comparison among different KL estimation methods.} All methods with non-negative output can achieve increased performance. The clipping method yields the best results, as it controls the magnitude of the KL value, leading to finer-grained updates and improved stability. The slightly lower result of forward KL proves the mode-seeking advantage of reverse KL.}
  \label{fig:ablation_kl}
  \vspace{-5pt}
\end{figure*}

\paragraph{Different KL Estimation Methods} Considering the large output space of VLMs, KL divergence cannot be computed analytically from logits; instead needs sampling. Consequently, the choice of the KL estimation method is critical, especially in GTR-Turbo. Directly computing the log-probability difference can yield negative estimates, which is problematic because we use the negative sentence-level KL as auxiliary rewards. As demonstrated by the blue curve in Figure \ref{fig:ablation_kl}, such KL estimation can grow increasingly negative, pushing the agent further away from the teacher.

Several approaches can resolve this issue. The simplest method is to clip the negative part or take its absolute value. Additionally, a non-negative and unbiased K3 estimator is proposed by Schulman \cite{schulmankldiv}. We also try the forward KL calculation. All estimators are presented in Equation \ref{eqn:kl_estimation}:

\vspace{-5pt}
\begin{align}
\label{eqn:kl_estimation}
    &\text{K1} = \log\pi_\theta - \log\pi_\text{merged}; & \nonumber \\
    &\text{KL}_\text{clip} = \text{clip}(\text{K1}, 0, +\infty); \nonumber \\
    &\text{KL}_\text{abs} = \lvert\text{K1}\rvert; \nonumber \\
    &\text{K3} = \text{K1} + e^{-\text{K1}} - 1; \nonumber \\
    &\text{KL}_\text{forward} = \text{clip}(-\text{K1}, 0, +\infty).
\end{align}
\vspace{-5pt}

Results in Figure \ref{fig:ablation_kl} demonstrate that all non-negative estimators lead to model improvements, the clipping method achieved the best results, and the differences among other methods are minor. As observed from the KL estimation curves, this is likely because clipping controls the scale of KL values, providing finer-grained updates and better stability when both the teacher and student are dynamically changing. 

The forward KL can also achieve high peak performance, but it still underperforms the reverse KL, consistent with prior studies \cite{jangrevkl, guminillm}. This is because reverse KL exhibits a ``mode-seeking'' characteristic, allowing the agent to capture a specific peak in the teacher's behavior rather than span over the broader distribution, making it more targeted and effective for guidance.

\begin{figure}[t]
    \centering
    \includegraphics[width=0.9\linewidth]{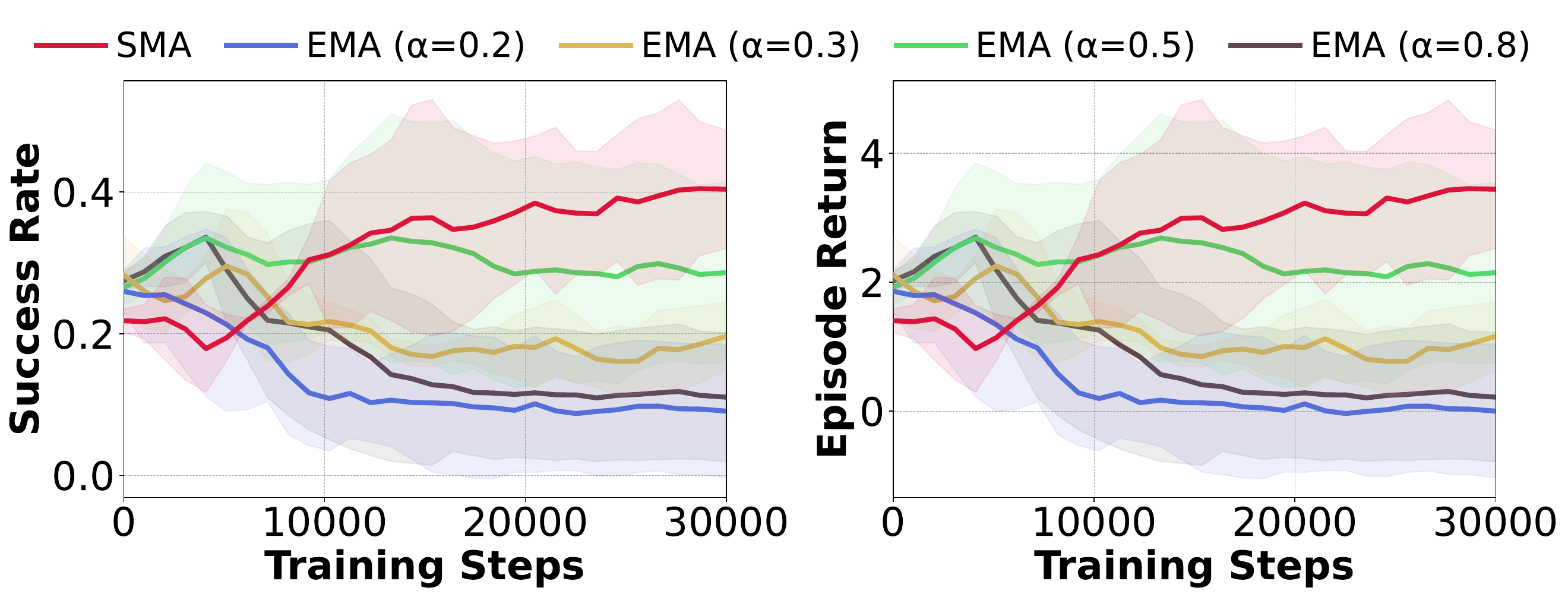}
    \caption{\textbf{Comparing different weights assignment methods.} Simple SMA already yields strong performance. A balanced choice of $\alpha$ is critical for realizing the benefit of EMA.}
    \vspace{-5pt}
    \label{fig:ablation_weight}  
\end{figure}

\paragraph{Different Weights Assignment Methods} As described in Section \ref{sec:model_merging}, there are different strategies for assigning weights during merging. In Figure \ref{fig:ablation_weight}, we explore various weight assignment methods and parameter choices.

The aforementioned experiments use the simplest arithmetic mean (SMA), which already yields satisfactory performance. The exponential moving average (EMA) strategy, controlled by a parameter $\alpha$, gives greater influence to more recent models with larger weights (see Equation \ref{eqn:weight_variants}). Results show that a balanced $\alpha=0.5$ performs the best among all candidates, with peak performance comparable to SMA.

An excessively high $\alpha$ causes the influence of historical checkpoints to diminish rapidly, weakening the benefits of model merging in terms of optimization smoothing and past knowledge integration. A very low $\alpha$, while theoretically closer to the SMA and shown to perform similarly in related pre-training research \cite{li2025model}, quickly fails in our experiments. 

Although a lower $\alpha$ results in a set of weights closer to average after convergence, the online recursive computation (Eqn. \ref{eqn:weight_variants}) introduces substantial bias when $k$ is small, leading to unbalanced merging where the latest models have minimal impact, thus degrading in the early stage. This issue is less pronounced in pretraining, where the number of checkpoints is much larger. Overall, choosing a balanced $\alpha$ is important for maximizing the effectiveness of the EMA strategy.

%% file: sec/5_conclusion.tex
\section{Conclusions and Limitations}
\label{sec:conclusion}
Credit assignment with sparse rewards remains a core challenge in the multi-turn RL of VLM agents. Previous process-guided approaches, such as GTR, rely on costly and possibly inaccessible external API models, severely limiting their scalability and practicality. In this work, we introduce the GTR-Turbo framework, which leverages the merged checkpoint as a free teacher and provides thought guidance through either SFT or KL-regularized objectives. This simple yet powerful framework enables self-evolving agents while reducing training time and cost, and achieving comparable or even superior performance to GTR. By better unleashing the model's decision-making and reflective capabilities, GTR-Turbo offers a more practical and efficient paradigm for complex multi-turn visual agentic tasks.

GTR-Turbo can be regarded as a bootstrapped training paradigm, which is feasible when external teacher models are inaccessible, or training data cannot be uploaded due to privacy concerns. However, in certain extreme scenarios in which the base model is too weak (e.g., an initial success rate of $<$5\%), self-improvement approaches can fail. Then traditional GTR is needed since an expensive but stronger external teacher can provide better guidance. Moreover, due to resource constraints, our experiments are primarily conducted on 7B models. Future research will verify the effectiveness of GTR-Turbo at different model sizes.

%% file: sec/X_suppl.tex
\onecolumn
\clearpage
\appendix

\section{Pseudocodes}
We present the GTR-Turbo pseudocodes, both for the SFT and KL thought guidance variants. 

\begin{algorithm}[htbp]
\caption{Training Procedure of GTR-Turbo (SFT)}
\label{alg:gtr}
	\begin{algorithmic}[1]
	    \State \textbf{Input:~}Environment 
        $\mathtt{env}$, agent model $\pi_{\theta_0}$, Replay buffer size $B$, update epoch $K$
            \State $\mathcal{C}\leftarrow[\pi_{\theta_0}]$ \Comment{Checkpoint buffer}
            \State $\mathcal{D}\leftarrow\varnothing$ \Comment{Thought dataset}
            \For{$k=0$ to $K-1$}
                \State $\mathcal{B}\leftarrow\varnothing$ \Comment{On-policy RL data buffer}
                \State Obtain $\pi_\text{merged}^{(k)}$ by merging all checkpoints in $\mathcal{C}$ \Comment{Eqn. 3}
                \State $o_t = \mathtt{env}$.reset()
                \While{$|\mathcal{B}| < B$}
                    \State Generate $(th_t, a_t)$ using $\pi_{\theta_k}$ given $o_t$
                    \State Generate $(\hat{th}_t, \hat{a}_t)$ using $\pi_\text{merged}^{(k)}$ given $o_t$ \Comment{Reference thought}
                    \State $r_t, o_{t+1} = \mathtt{env}$.step($a_t$)
                    \State $\mathcal{B} \leftarrow \mathcal{B} \cup (o_t, a_t, r_t, o_{t+1})$
                    \State $\mathcal{D} \leftarrow \mathcal{D} \cup (o_t, \hat{th}_t)$
                \EndWhile
                \State Sample mini-batch $b$ from $\mathcal{B}$, $d$ from $\mathcal{D}$
                \State Compute $\mathcal{L}_{\text{PPO}}$ with $b$
                \State Compute $\mathcal{L}_{\text{SFT}}$ with $d$ \Comment{Eqn. 2}
                \State $\theta_{k+1} = \arg\min_\theta(\mathcal{L}_{\text{PPO}} + \mathcal{L}_{\text{SFT}})$ \Comment{Eqn. 5}
                \State $\mathcal{C} \leftarrow \mathcal{C} \cup \pi_{\theta_{k+1}}$
            \EndFor
            \State \textbf{Output:~}$\pi_{\theta_K}$
	\end{algorithmic}
\end{algorithm}

\begin{algorithm}[htbp]
\caption{Training Procedure of GTR-Turbo (KL)}
\label{alg:gtr}
	\begin{algorithmic}[1]
	    \State \textbf{Input:~}Environment 
        $\mathtt{env}$, agent model $\pi_{\theta_0}$, Replay buffer size $B$, update epoch $K$
            \State $\mathcal{C}\leftarrow[\pi_{\theta_0}]$ \Comment{Checkpoint buffer}
            \For{$k=0$ to $K-1$}
                \State $\mathcal{B}\leftarrow\varnothing$ \Comment{On-policy RL data buffer}
                \State Obtain $\pi_\text{merged}^{(k)}$ by merging all checkpoints in $\mathcal{C}$ \Comment{Eqn.3}
                \State $o_t = \mathtt{env}$.reset()
                \While{$|\mathcal{B}| < B$}
                    \State Generate $(th_t, a_t)$ using $\pi_{\theta_k}$ given $o_t$
                    \State Calculate $\text{RevKL}\left(\pi_{\theta_k}, \pi_\text{merged}^{(k)};th_t\right)$ \Comment{Eqn. 6}
                    \State $r_t, o_{t+1} = \mathtt{env}$.step($a_t$)
                    \State $\mathcal{B} \leftarrow \mathcal{B} \cup \left(o_t, a_t, r_t- \beta \cdot\text{RevKL}\left(\pi_{\theta_k}, \pi_\text{merged}^{(k)};th_t\right), o_{t+1}\right)$
                \EndWhile
                \State Sample mini-batch $b$ from $\mathcal{B}$
                \State Compute $\mathcal{L}_{\text{PPO}}$ with $b$
                \State $\theta_{k+1} = \arg\min_\theta\mathcal{L}_{\text{PPO}}$
                \State $\mathcal{C} \leftarrow \mathcal{C} \cup \pi_{\theta_{k+1}}$
            \EndFor
            \State \textbf{Output:~}$\pi_{\theta_K}$
	\end{algorithmic}
\end{algorithm}

In GTR-Turbo (KL), $\beta$ controls the contribution of the reserve KL term within the reward. Throughout this paper, we use the default setting of $\beta=1$.

\section{Additional Details on Training}
\subsection{Training Setting}
Drawing inspiration from the common practice in RL post-training frameworks, \cite{ouyang2022training, zhai2025fine, wei2025gtr}, we perform one epoch of supervised fine-tuning on the base Qwen2.5-VL model \cite{bai2025qwen2.5} before RL training, so that the agent possesses a basic instruction-following capability. The datasets are sourced from the RL4VLM paper \cite{zhai2025fine}, with labels for the Points24 provided by a task solver and labels for the ALFWorld environment generated by GPT-4V. 

\subsection{Hyperparameters}
We provide the hyperparameters used for GTR-Turbo training in Table \ref{tab:hyperparameters}, which are primarily derived from previous work \cite{zhai2025fine, wei2025gtr}. We employ LoRA \cite{hu2022lora} to fine-tune the entire VLM model.

\begin{table}[htb]
\caption{\textbf{Hyperparameters of GTR-Turbo}}
\label{tab:hyperparameters}
\begin{center}
\begin{small}
\begin{tabular}{lcc}
\toprule
\textbf{Hyperparameter} & & \textbf{Value}  \\
\midrule
\multicolumn{3}{c}{\textbf{General Setup - Training}} \\
Learning rate     && $\mathtt{CosineAnnealingLR}$ \\
Initial learning rate     && $1e-5$ \\
Final learning rate    && $1e-9$ \\
Maximum learning rate step  && $25$ \\
Discount factor $\gamma$  && $0.9$ \\
GAE $\lambda$ && $0.95$ \\
PPO entropy coefficient && $0.01$ \\
PPO value loss coefficient && $0.5$ \\
PPO clip parameter $c$ && $0.1$ \\
PPO epoch && $4$ \\
Gradient accumulation steps && $128$ \\
LoRA $r$ && $128$ \\
LoRA $\alpha$ && $256$ \\
LoRA $\mathtt{dropout}$ && $0.05$ \\
KL loss coefficient $\beta$ & (for KL guidance) & $1$ \\
\midrule
\multicolumn{3}{c}{\textbf{General Setup - Models}} \\
Generation max text length && $256$ \\
Generation temperature && $0.2$ \\
Generation repetition penalty && $1.2$ \\
Model Merging Method  && TIES \\
TIES Density  && 0.8 \\
Teacher Generation base temperature & (for SFT guidance)  & $0.2$ \\
Teacher Generation max temperature  & (for SFT guidance)  & $0.9$ \\
Teacher Generation temperature retry coefficient & (for SFT guidance)  & $1.1$ \\
\midrule
\multicolumn{3}{c}{\textbf{For Points24 task}} \\
Environmental steps && $30000$ \\
Thought probability coefficient && $0.5$ \\
\midrule
\multicolumn{3}{c}{\textbf{For ALFWorld task}} \\
Environmental steps && $20000$ \\
Thought probability coefficient && $0.2$ \\
\bottomrule
\end{tabular}
\end{small}
\end{center}
\end{table}

\newpage
\section{Additional Experiment Results}
\subsection{Results on stronger and more recent models}
We also evaluate the efficacy of GTR-Turbo using the newly released Qwen3-VL-8B-Instruct model. We evaluate on ALFWorld using the KL variant of GTR-Turbo. The results show that GTR-Turbo remains compatible with the latest model family, and the stronger base capability of Qwen3-VL leads to improved performance, even surpassing the success rate of Qwen2.5-VL-32B, a model that is four times larger in scale.

Moreover, we observe that, in general-knowledge reasoning tasks such as ALFWorld, Qwen3-VL can perform RL directly without any SFT initialization. This suggests that as foundation models continue to evolve, GTR-Turbo may become even simpler to use and more broadly applicable.

\begin{figure}[htbp]
  \centering
  \includegraphics[width=0.35\linewidth]{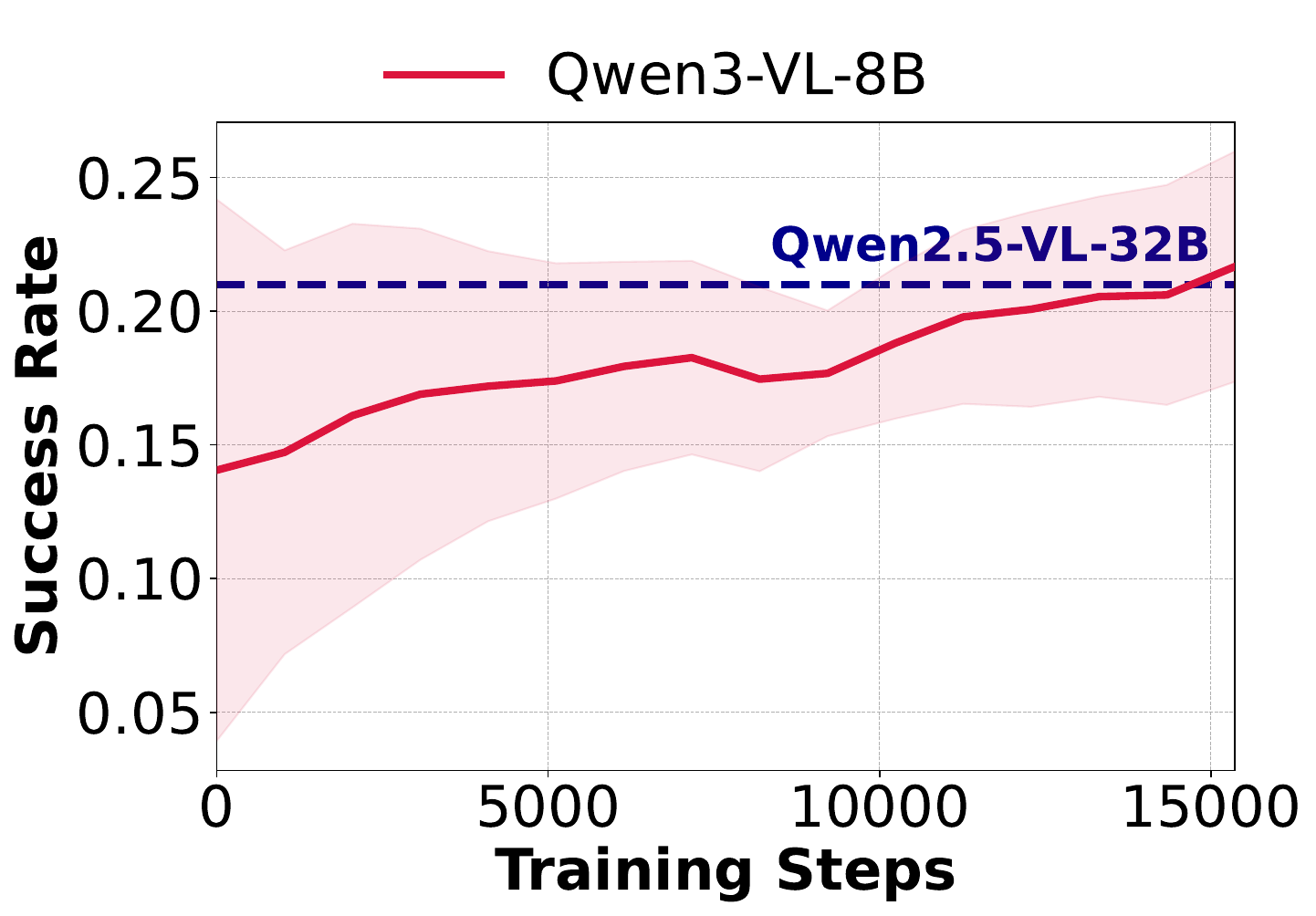}
  \caption{\textbf{Result of Qwen3-VL-8B on ALFWorld.}}
  \label{fig:qwen3_alfworld}
  \vspace{-5pt}
\end{figure}

\subsection{Additional Experiments on other open-ended tasks}
We conduct an experiment on the challenging GUI benchmark Android-in-the-Wild (AitW) \cite{rawles2023androidinthewild} using Qwen3-VL-8B-Instruct, as shown in Table~\ref{tab:supp_gui}. GTR-Turbo still outperforms strong DigiRL and PPO baselines without heavy hyperparameter tuning and reward shaping. Moreover, we also compare the quality of reasoning traces between PPO and GTR-Turbo using GPT-5.2 with a simple LLM-as-a-judge method. These results demonstrate the efficacy of GTR-Turbo across diverse visual environments.

\begin{table}[htb]
\centering
    \resizebox{0.4\linewidth}{!}{
    \begin{tabular}{ccc}\toprule
    \textbf{Method} & \textbf{Success Rate} & \textbf{Reasoning Score} \\
    \midrule
    DigiRL   & 71.9\% & -\\
    PPO    & 75.0\% & 3.26\\
    \textbf{GTR-Turbo}  & \textbf{80.2\%} & \textbf{3.93} \\ \bottomrule
    \end{tabular}
    }
    \caption{\textbf{Experiment results on Android-in-the-Wild.}}
    \label{tab:supp_gui}
\end{table}

\begin{figure*}[t]
\begin{floatrow}
\ffigbox[0.3\textwidth]{%
\centering
\includegraphics[width=\linewidth]{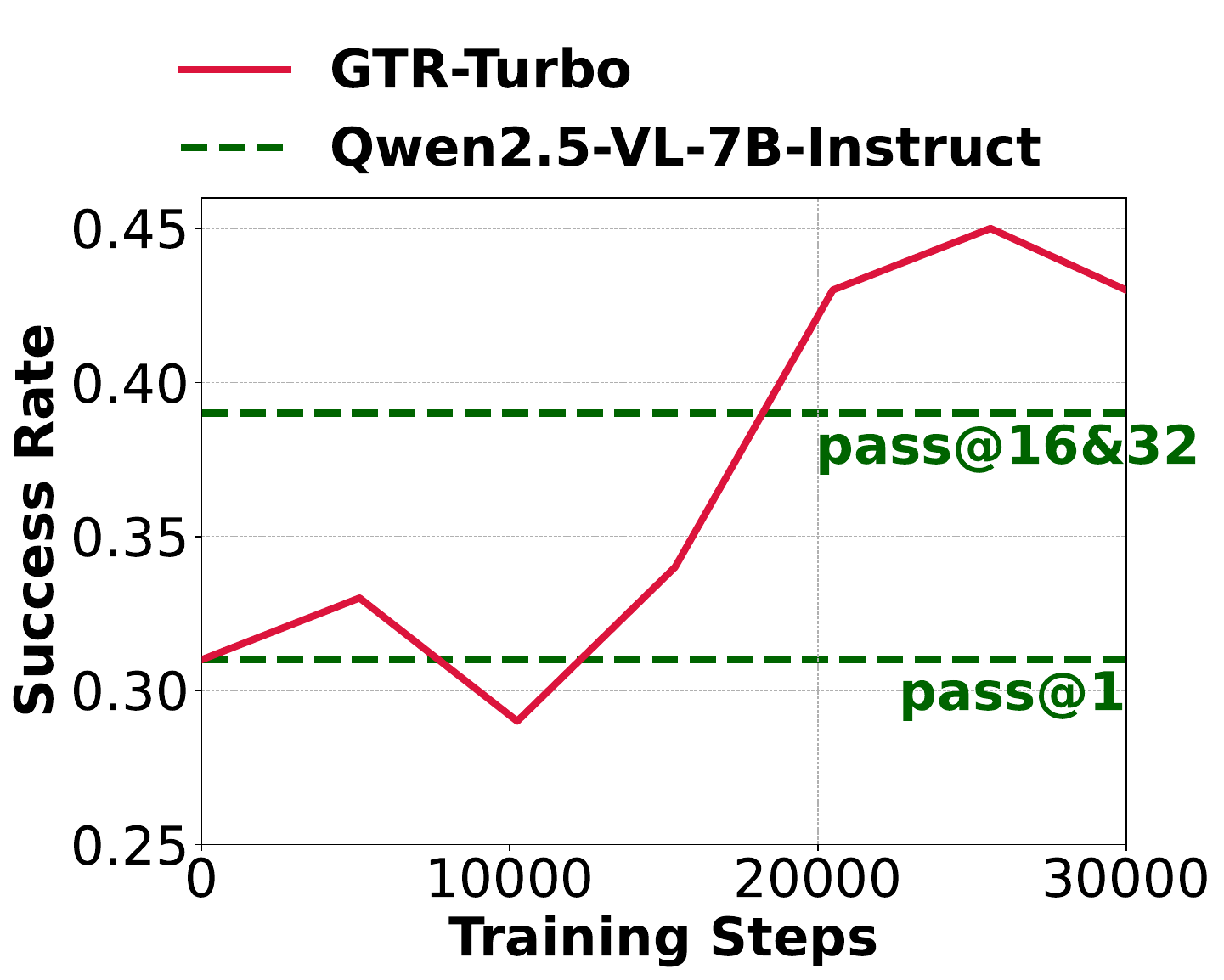}
}{%
  \caption{\textbf{Comparison of GTR-Turbo training curve with the pass@k results of the base model.}}
  \label{fig:supp_pass_k} 
}
\ffigbox[0.3\textwidth]{%
\includegraphics[width=\linewidth]{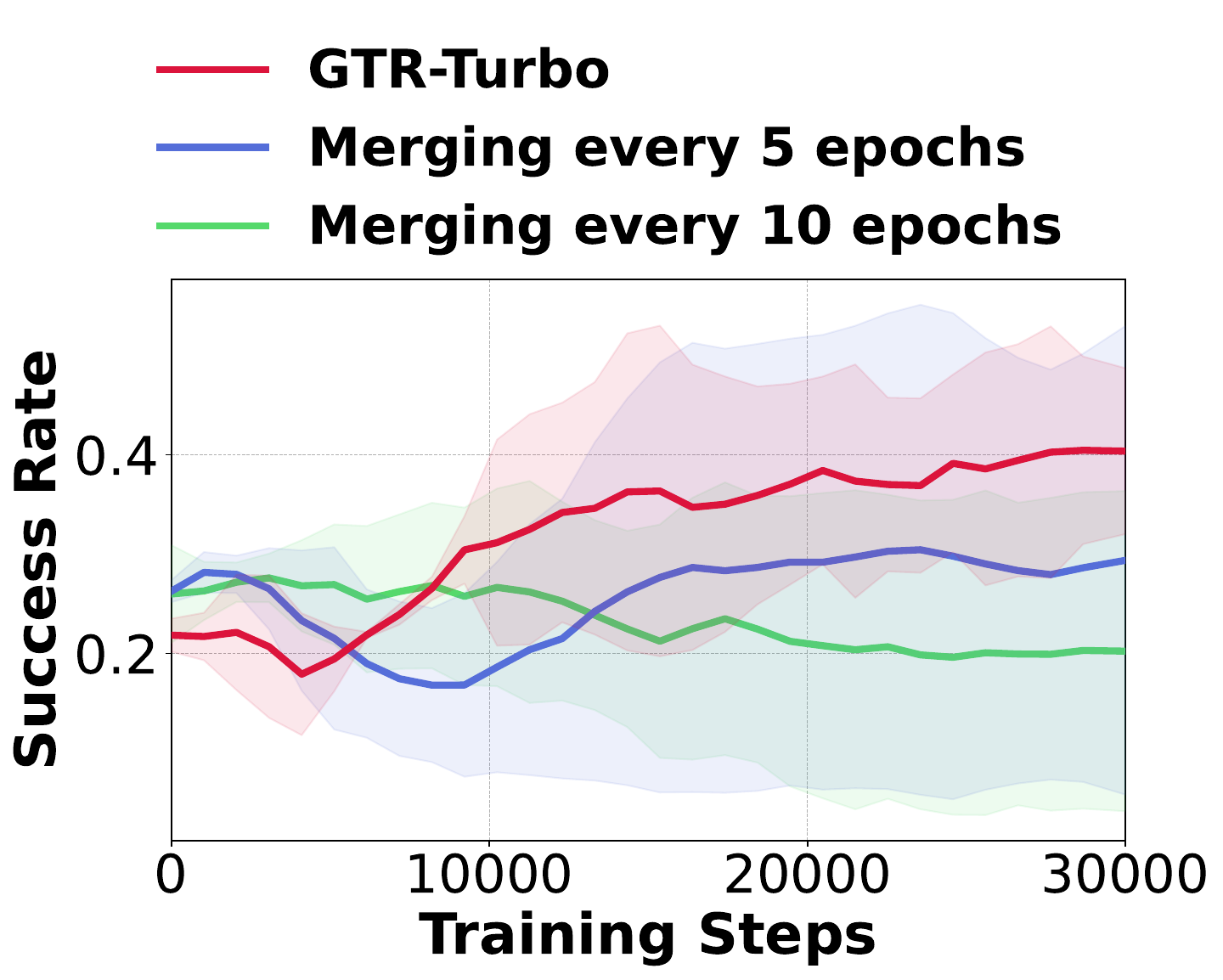}
}{%
  \caption{\textbf{Success rate results of GTR-Turbo using different merging frequencies.}}
  \label{fig:supp_freq} 
}
\hspace{10pt}
\ffigbox[0.32\textwidth]{%
\includegraphics[width=\linewidth]{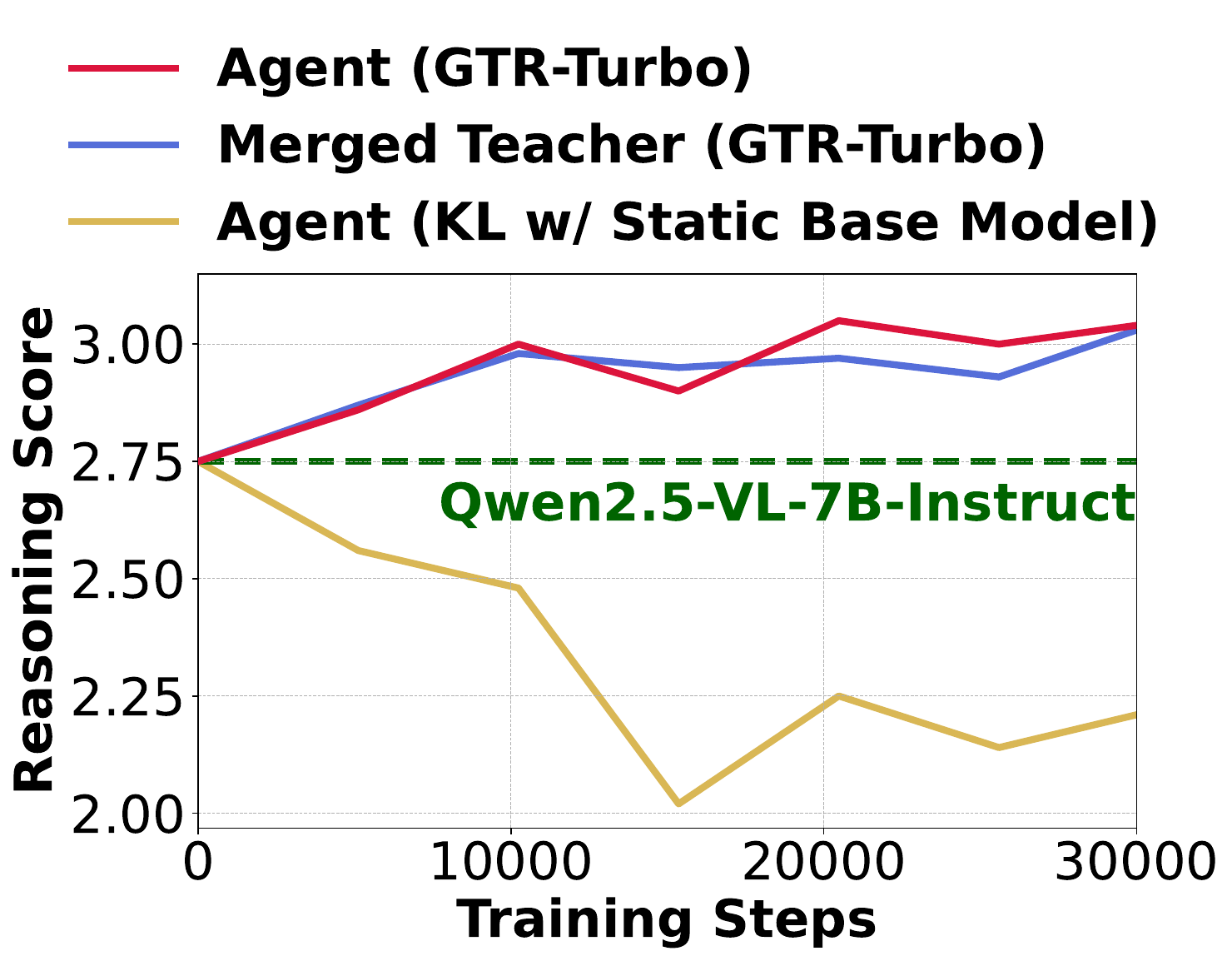}
}{%
  \caption{\textbf{Reasoning score evaluation of models in GTR-Turbo and baseline with static teacher.}}
  \label{fig:supp_reasoning} 
}
\end{floatrow}
\vspace{-12pt}
\end{figure*}

\subsection{Pass@k Comparison}
Following prior work \cite{yue2025does}, we use $pass@k$ success rate with increasing $k$ until convergence ($ k=32$ achieves the same performance as $ k=16$) to assess the upper bound of the base model's capability.
Figure \ref{fig:supp_pass_k} shows that the agent trained by GTR-Turbo can easily surpass the ceiling, indicating that GTR-Turbo enables the model to acquire capabilities beyond its original distribution.

\subsection{Ablation study regarding merging frequency}
In Figure \ref{fig:supp_freq}, we ablate the merging frequency. GTR-Turbo continues to yield appealing results up to a merging interval of 10, demonstrating its robustness to this hyperparameter.

\subsection{Reasoning Score Evaluation}
To further verify whether GTR-Turbo can mitigate ``thought collapse'', we employ an LLM-as-a-judge approach using GPT-5.2 to evaluate the quality of agent-generated reasoning traces during RL in terms of factual correctness, logical rigor, and coherence. Figure \ref{fig:supp_reasoning} shows that the RL-trained agent without guidance from the merged teacher exhibits a clear ``thought collapse'' phenomenon. This indicates that the merged checkpoint is certainly a teacher with better reasoning rather than a simple regularized reference.

\section{Additional Details on Environments}
We provide a detailed introduction to the experimental environments used in this study.

\subsection{Points24}
\paragraph{State and action space.} At each observation $o_t$ in the Points24 task, the agent observes an image showing four poker cards and a text-based representation of the current formula. The goal is to form a formula equal to 24 using the numbers represented by the four cards and basic operators. Cards ``J'', ``Q'', ``K'' are all treated as number 10. The action space includes \{``1'', ``2'', $\ldots$, ``10'', ``+'', ``-'', ``*'', ``/'', ``('', ``)'', ``=''\}, and each card can only be used once. Selecting a number not present in the image or one that has already been used is considered an illegal action. If the action is legal, the corresponding number or operator is appended to the current formula, forming the next observation $o_{t+1}$; if the action is illegal, the state remains unchanged $o_{t+1}=o_t$. The environment does not guarantee that the four cards in the image have a feasible solution equal to 24.

\paragraph{Reward function.} At each step, the agent receives a reward $r=-1$ for outputting an illegal action and a reward $r=0$ for a legal action. The episode terminates when the agent outputs ``='' as an action or the step count exceeds $T=20$. At termination, if the formula evaluates to 24, the agent receives an outcome reward $r=10$; otherwise, it receives $r=-1$.

\begin{figure}[ht]
  \centering
    \includegraphics[width=0.75\linewidth]{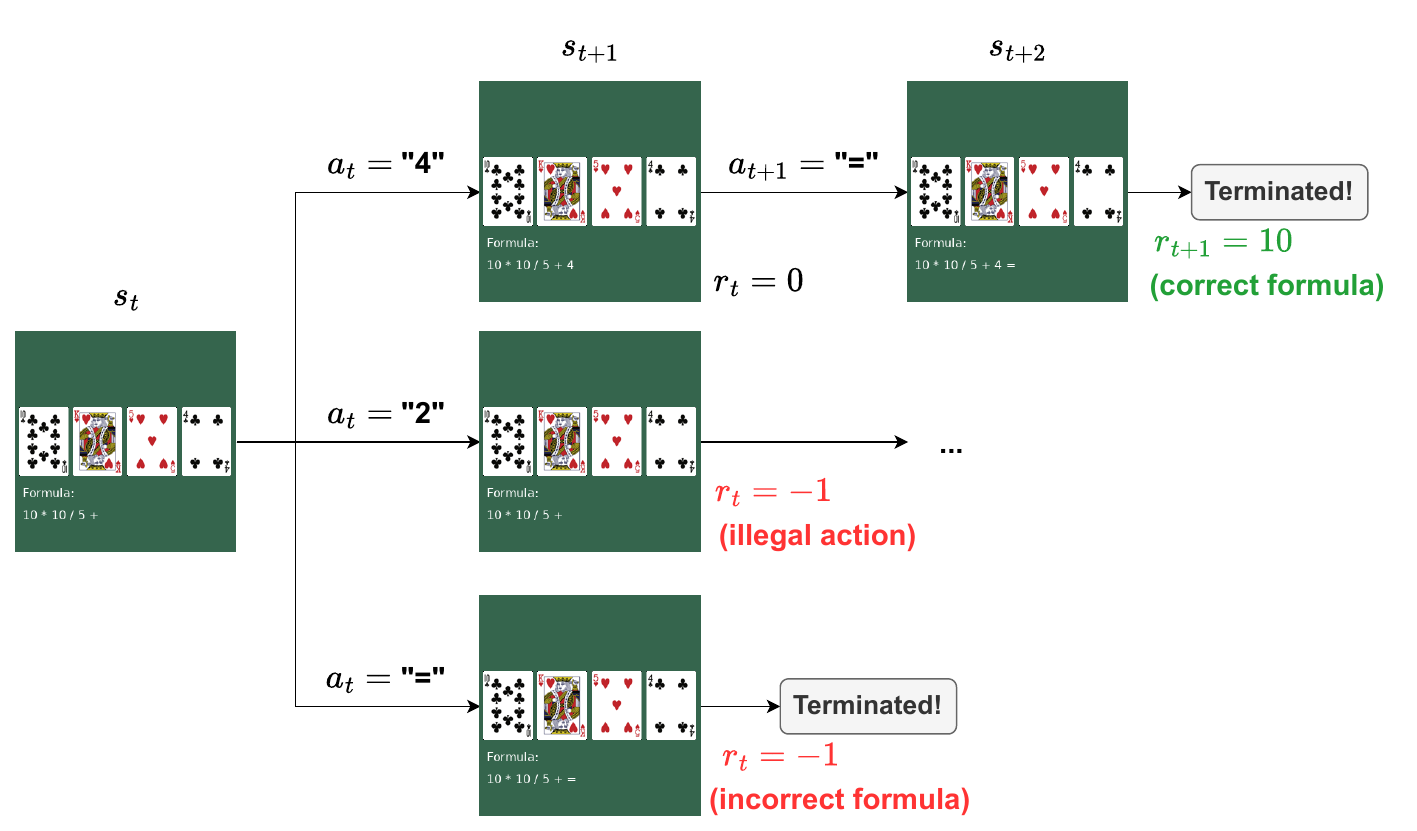}
  \caption{\textbf{The Points24 task.}}   
\end{figure}

\subsection{ALFWorld}
\paragraph{State and action space.} In the ALFWorld environment in our experiments, the agent receives an RGB observation image and a history of past actions at each observation $o_{t}$. The action space includes all possible interactions in the current scenario, typically categorized as: (1) $\mathtt{go\ to\ \{recep\}}$, (2) $\mathtt{take\ \{obj\} \ from\ \{recep\}}$, (3) $\mathtt{put \ \{obj\} \ in/on\ \{recep\}}$, (4) $\mathtt{open\ \{recep\}}$, (5) $\mathtt{close\ \{recep\}}$, (6) $\mathtt{toggle\ \{obj\}\ \{recep\}}$, (7) $\mathtt{clean\ \{obj\} \ with\ \{recep\}}$, (8) $\mathtt{heat\ \{obj\} \ with\ \{recep\}}$, (9) $\mathtt{cool\ \{obj\} \ with\ \{recep\}}$, where $\mathtt{\{obj\}}$ and $\mathtt{\{recep\}}$ denote objects and receptacles. After an admissible action is taken, ALFWorld renders the updated scene from the agent's view as the next observation $o_{t+1}$. $o_{t+1}=o_t$ if the action is illegal.

Notably, the ALFWorld environment provides both an image and a text description of the observation scene at each step. As noted in GTR, the VLM agent may rely heavily on textual descriptions rather than visual observations, which contradicts the purpose of visual agentic tasks. GTR therefore modified the state by removing the text description, which we adopt in GTR-Turbo. We also align with GTR by including the action history in the input prompt to more closely simulate real-world scenarios. These adjustments increase the task's difficulty, thereby emphasizing the agent's comprehensive visual recognition and long-horizon decision-making capabilities.

\paragraph{Reward function.} The reward of ALFWorld consists of two components. Each observation $o$ has a set of admissible actions $\mathcal{A}_\text{adm}(s)$, and illegal actions are penalized. Additionally, each task in ALFWorld has both the final goal $g_\text{task}$ and sub-goals $g_\text{sub}$, and achieving these goals also provides rewards. Formally, the reward function can be written as:
\begin{equation}
    r(s_t, a_t, s_{t+1}|g_\text{task}) = 50 \times \mathbf{1}(s_{t+1} = g_\text{task}) + \mathbf{1}(s_{t+1} = g_\text{sub}) - \mathbf{1}(a_t \notin \mathcal{A}_\text{adm}(s)).
\end{equation}

\begin{figure}[ht]
  \centering
    \includegraphics[width=\linewidth]{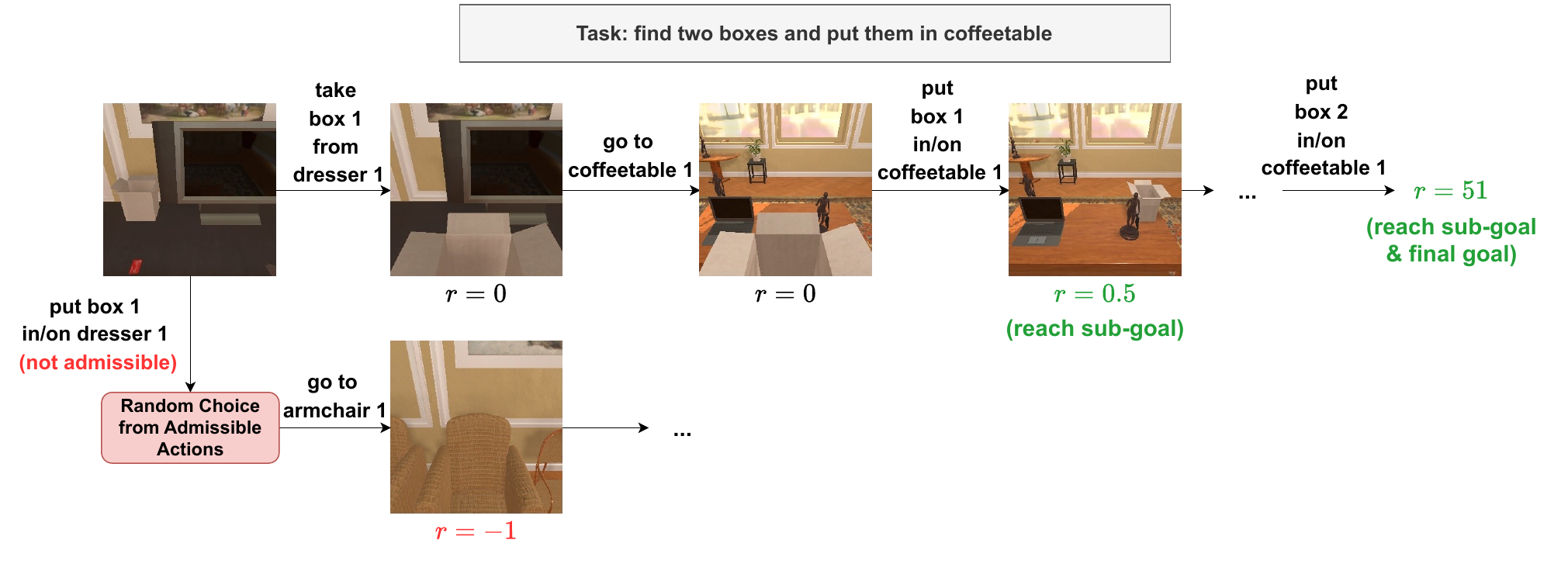}
  \caption{\textbf{The ALFWorld task.}}   
\end{figure}

%% file: main.bib
@String(ICLR = {Int. Conf. Learn. Represent.})

@String(ICLR  = {ICLR})

@article{wei2022chain,
  title={Chain-of-thought prompting elicits reasoning in large language models},
  author={Wei, Jason and Wang, Xuezhi and Schuurmans, Dale and Bosma, Maarten and Xia, Fei and Chi, Ed and Le, Quoc V and Zhou, Denny and others},
  journal={Advances in neural information processing systems},
  volume={35},
  pages={24824--24837},
  year={2022}
}

@article{yao2023tree,
  title={Tree of thoughts: Deliberate problem solving with large language models},
  author={Yao, Shunyu and Yu, Dian and Zhao, Jeffrey and Shafran, Izhak and Griffiths, Tom and Cao, Yuan and Narasimhan, Karthik},
  journal={Advances in neural information processing systems},
  volume={36},
  pages={11809--11822},
  year={2023}
}

@inproceedings{yao2023react,
  title={React: Synergizing reasoning and acting in language models},
  author={Yao, Shunyu and Zhao, Jeffrey and Yu, Dian and Du, Nan and Shafran, Izhak and Narasimhan, Karthik and Cao, Yuan},
  booktitle={International conference on learning representations},
  year={2023}
}

@article{wang2023describe,
  title={Describe, explain, plan and select: Interactive planning with large language models enables open-world multi-task agents},
  author={Wang, Zihao and Cai, Shaofei and Chen, Guanzhou and Liu, Anji and Ma, Xiaojian and Liang, Yitao},
  journal={arXiv preprint arXiv:2302.01560},
  year={2023}
}

@inproceedings{lightman2023let,
  title={Let's verify step by step},
  author={Lightman, Hunter and Kosaraju, Vineet and Burda, Yuri and Edwards, Harrison and Baker, Bowen and Lee, Teddy and Leike, Jan and Schulman, John and Sutskever, Ilya and Cobbe, Karl},
  booktitle={The twelfth international conference on learning representations},
  year={2023}
}

@article{wang2023voyager,
  title={Voyager: An open-ended embodied agent with large language models},
  author={Wang, Guanzhi and Xie, Yuqi and Jiang, Yunfan and Mandlekar, Ajay and Xiao, Chaowei and Zhu, Yuke and Fan, Linxi and Anandkumar, Anima},
  journal={arXiv preprint arXiv:2305.16291},
  year={2023}
}

@inproceedings{park2023generative,
  title={Generative agents: Interactive simulacra of human behavior},
  author={Park, Joon Sung and O'Brien, Joseph and Cai, Carrie Jun and Morris, Meredith Ringel and Liang, Percy and Bernstein, Michael S},
  booktitle={Proceedings of the 36th annual acm symposium on user interface software and technology},
  pages={1--22},
  year={2023}
}

@article{huang2023voxposer,
  title={Voxposer: Composable 3d value maps for robotic manipulation with language models},
  author={Huang, Wenlong and Wang, Chen and Zhang, Ruohan and Li, Yunzhu and Wu, Jiajun and Fei-Fei, Li},
  journal={arXiv preprint arXiv:2307.05973},
  year={2023}
}

@article{ahn2022can,
  title={Do as i can, not as i say: Grounding language in robotic affordances},
  author={Ahn, Michael and Brohan, Anthony and Brown, Noah and Chebotar, Yevgen and Cortes, Omar and David, Byron and Finn, Chelsea and Fu, Chuyuan and Gopalakrishnan, Keerthana and Hausman, Karol and others},
  journal={arXiv preprint arXiv:2204.01691},
  year={2022}
}

@inproceedings{huang2022language,
  title={Language models as zero-shot planners: Extracting actionable knowledge for embodied agents},
  author={Huang, Wenlong and Abbeel, Pieter and Pathak, Deepak and Mordatch, Igor},
  booktitle={International conference on machine learning},
  pages={9118--9147},
  year={2022},
  organization={PMLR}
}

@article{shinn2023reflexion,
  title={Reflexion: Language agents with verbal reinforcement learning},
  author={Shinn, Noah and Cassano, Federico and Gopinath, Ashwin and Narasimhan, Karthik and Yao, Shunyu},
  journal={Advances in neural information processing systems},
  volume={36},
  pages={8634--8652},
  year={2023}
}

@article{wu2023autogen,
  title={Autogen: Enabling next-gen llm applications via multi-agent conversation},
  author={Wu, Qingyun and Bansal, Gagan and Zhang, Jieyu and Wu, Yiran and Li, Beibin and Zhu, Erkang and Jiang, Li and Zhang, Xiaoyun and Zhang, Shaokun and Liu, Jiale and others},
  journal={arXiv preprint arXiv:2308.08155},
  year={2023}
}

@article{driess2023palm,
  title={Palm-e: An embodied multimodal language model},
  author={Driess, Danny and Xia, Fei and Sajjadi, Mehdi SM and Lynch, Corey and Chowdhery, Aakanksha and Wahid, Ayzaan and Tompson, Jonathan and Vuong, Quan and Yu, Tianhe and Huang, Wenlong and others},
    journal={International conference on machine learning},
    year={2023}
}

@inproceedings{gao2024physically,
  title={Physically grounded vision-language models for robotic manipulation},
  author={Gao, Jensen and Sarkar, Bidipta and Xia, Fei and Xiao, Ted and Wu, Jiajun and Ichter, Brian and Majumdar, Anirudha and Sadigh, Dorsa},
  booktitle={2024 IEEE international conference on robotics and automation},
  pages={12462--12469},
  year={2024},
  organization={IEEE}
}

@article{mu2023embodiedgpt,
  title={Embodiedgpt: Vision-language pre-training via embodied chain of thought},
  author={Mu, Yao and Zhang, Qinglong and Hu, Mengkang and Wang, Wenhai and Ding, Mingyu and Jin, Jun and Wang, Bin and Dai, Jifeng and Qiao, Yu and Luo, Ping},
  journal={Advances in neural information processing systems},
  volume={36},
  pages={25081--25094},
  year={2023}
}

@article{sumers2023distilling,
  title={Distilling internet-scale vision-language models into embodied agents},
  author={Sumers, Theodore and Marino, Kenneth and Ahuja, Arun and Fergus, Rob and Dasgupta, Ishita},
  journal={arXiv preprint arXiv:2301.12507},
  year={2023}
}

@inproceedings{yang2024octopus,
  title={Octopus: Embodied vision-language programmer from environmental feedback},
  author={Yang, Jingkang and Dong, Yuhao and Liu, Shuai and Li, Bo and Wang, Ziyue and Tan, Haoran and Jiang, Chencheng and Kang, Jiamu and Zhang, Yuanhan and Zhou, Kaiyang and others},
  booktitle={European conference on computer vision},
  pages={20--38},
  year={2024},
  organization={Springer}
}

@article{brohan2023rt,
  title={Rt-2: Vision-language-action models transfer web knowledge to robotic control},
  author={Brohan, Anthony and Brown, Noah and Carbajal, Justice and Chebotar, Yevgen and Chen, Xi and Choromanski, Krzysztof and Ding, Tianli and Driess, Danny and Dubey, Avinava and Finn, Chelsea and others},
  journal={arXiv preprint arXiv:2307.15818},
  year={2023}
}

@article{ouyang2022training,
  title={Training language models to follow instructions with human feedback},
  author={Ouyang, Long and Wu, Jeffrey and Jiang, Xu and Almeida, Diogo and Wainwright, Carroll and Mishkin, Pamela and Zhang, Chong and Agarwal, Sandhini and Slama, Katarina and Ray, Alex and others},
  journal={Advances in neural information processing systems},
  volume={35},
  pages={27730--27744},
  year={2022}
}

@article{uesato2022solving,
  title={Solving math word problems with process-and outcome-based feedback},
  author={Uesato, Jonathan and Kushman, Nate and Kumar, Ramana and Song, Francis and Siegel, Noah and Wang, Lisa and Creswell, Antonia and Irving, Geoffrey and Higgins, Irina},
  journal={arXiv preprint arXiv:2211.14275},
  year={2022}
}

@article{shao2024deepseekmath,
  title={Deepseekmath: Pushing the limits of mathematical reasoning in open language models},
  author={Shao, Zhihong and Wang, Peiyi and Zhu, Qihao and Xu, Runxin and Song, Junxiao and Bi, Xiao and Zhang, Haowei and Zhang, Mingchuan and Li, YK and Wu, Y and others},
  journal={arXiv preprint arXiv:2402.03300},
  year={2024}
}

@article{zhai2025fine,
  title={Fine-tuning large vision-language models as decision-making agents via reinforcement learning},
  author={Zhai, Simon and Bai, Hao and Lin, Zipeng and Pan, Jiayi and Tong, Peter and Zhou, Yifei and Suhr, Alane and Xie, Saining and LeCun, Yann and Ma, Yi and others},
  journal={Advances in neural information processing systems},
  volume={37},
  pages={110935--110971},
  year={2025}
}

@article{wang2024q,
  title={Q*: Improving multi-step reasoning for llms with deliberative planning},
  author={Wang, Chaojie and Deng, Yanchen and Lyu, Zhiyi and Zeng, Liang and He, Jujie and Yan, Shuicheng and An, Bo},
  journal={arXiv preprint arXiv:2406.14283},
  year={2024}
}

@article{cui2025process,
  title={Process reinforcement through implicit rewards},
  author={Cui, Ganqu and Yuan, Lifan and Wang, Zefan and Wang, Hanbin and Li, Wendi and He, Bingxiang and Fan, Yuchen and Yu, Tianyu and Xu, Qixin and Chen, Weize and others},
  journal={arXiv preprint arXiv:2502.01456},
  year={2025}
}

@article{yuan2024free,
  title={Free process rewards without process labels},
  author={Yuan, Lifan and Li, Wendi and Chen, Huayu and Cui, Ganqu and Ding, Ning and Zhang, Kaiyan and Zhou, Bowen and Liu, Zhiyuan and Peng, Hao},
  journal={arXiv preprint arXiv:2412.01981},
  year={2024}
}

@article{zhang2024generative,
  title={Generative verifiers: Reward modeling as next-token prediction},
  author={Zhang, Lunjun and Hosseini, Arian and Bansal, Hritik and Kazemi, Mehran and Kumar, Aviral and Agarwal, Rishabh},
  journal={arXiv preprint arXiv:2408.15240},
  year={2024}
}

@article{gao2024llm,
  title={Llm critics help catch bugs in mathematics: Towards a better mathematical verifier with natural language feedback},
  author={Gao, Bofei and Cai, Zefan and Xu, Runxin and Wang, Peiyi and Zheng, Ce and Lin, Runji and Lu, Keming and Liu, Dayiheng and Zhou, Chang and Xiao, Wen and others},
  journal={arXiv preprint arXiv:2406.14024},
  year={2024}
}

@article{xia2024evaluating,
  title={Evaluating mathematical reasoning beyond accuracy},
  author={Xia, Shijie and Li, Xuefeng and Liu, Yixin and Wu, Tongshuang and Liu, Pengfei},
  journal={arXiv preprint arXiv:2404.05692},
  year={2024}
}

@article{guo2025deepseek,
  title={Deepseek-r1: Incentivizing reasoning capability in llms via reinforcement learning},
  author={Guo, Daya and Yang, Dejian and Zhang, Haowei and Song, Junxiao and Zhang, Ruoyu and Xu, Runxin and Zhu, Qihao and Ma, Shirong and Wang, Peiyi and Bi, Xiao and others},
  journal={arXiv preprint arXiv:2501.12948},
  year={2025}
}

@article{shridhar2020alfworld,
  title={Alfworld: Aligning text and embodied environments for interactive learning},
  author={Shridhar, Mohit and Yuan, Xingdi and C{\^o}t{\'e}, Marc-Alexandre and Bisk, Yonatan and Trischler, Adam and Hausknecht, Matthew},
  journal={arXiv preprint arXiv:2010.03768},
  year={2020}
}

@article{schulman2017proximal,
  title={Proximal policy optimization algorithms},
  author={Schulman, John and Wolski, Filip and Dhariwal, Prafulla and Radford, Alec and Klimov, Oleg},
  journal={arXiv preprint arXiv:1707.06347},
  year={2017}
}

@inproceedings{ross2011reduction,
  title={A reduction of imitation learning and structured prediction to no-regret online learning},
  author={Ross, St{\'e}phane and Gordon, Geoffrey and Bagnell, Drew},
  booktitle={Proceedings of the fourteenth international conference on artificial intelligence and statistics},
  pages={627--635},
  year={2011},
  organization={JMLR Workshop and Conference Proceedings}
}

@article{hu2022lora,
  title={Lora: Low-rank adaptation of large language models.},
  author={Hu, Edward J and Shen, Yelong and Wallis, Phillip and Allen-Zhu, Zeyuan and Li, Yuanzhi and Wang, Shean and Wang, Lu and Chen, Weizhu and others},
  journal={ICLR},
  volume={1},
  number={2},
  pages={3},
  year={2022}
}

@inproceedings{yang2024embodied,
  title={Embodied multi-modal agent trained by an llm from a parallel textworld},
  author={Yang, Yijun and Zhou, Tianyi and Li, Kanxue and Tao, Dapeng and Li, Lusong and Shen, Li and He, Xiaodong and Jiang, Jing and Shi, Yuhui},
  booktitle={Proceedings of the IEEE/CVF conference on computer vision and pattern recognition},
  pages={26275--26285},
  year={2024}
}

@article{yu2026proact,
  title={ProAct: Agentic Lookahead in Interactive Environments},
  author={Yu, Yangbin and Yang, Mingyu and Li, Junyou and Gao, Yiming and Liu, Feiyu and Yang, Yijun and Lin, Zichuan and Lyu, Jiafei and Liu, Yicheng and Lu, Zhicong and others},
  journal={arXiv preprint arXiv:2602.05327},
  year={2026}
}

@article{zhou2024wall,
  title={Wall-e: World alignment by rule learning improves world model-based llm agents},
  author={Zhou, Siyu and Zhou, Tianyi and Yang, Yijun and Long, Guodong and Ye, Deheng and Jiang, Jing and Zhang, Chengqi},
  journal={arXiv preprint arXiv:2410.07484},
  year={2024}
}

@article{wei2025gtr,
  title={GTR: Guided Thought Reinforcement Prevents Thought Collapse in RL-based VLM Agent Training},
  author={Wei, Tong and Yang, Yijun and Xing, Junliang and Shi, Yuanchun and Lu, Zongqing and Ye, Deheng},
  journal={arXiv preprint arXiv:2503.08525},
  year={2025}
}

@article{lu2025onpolicydistillation,
  author = {Kevin Lu and Thinking Machines Lab},
  title = {On-Policy Distillation},
  journal = {Thinking Machines Lab: Connectionism},
  year = {2025},
  note = {https://thinkingmachines.ai/blog/on-policy-distillation},
  doi = {10.64434/tml.20251026},
}

@article{wang2025ragen,
  title={Ragen: Understanding self-evolution in llm agents via multi-turn reinforcement learning},
  author={Wang, Zihan and Wang, Kangrui and Wang, Qineng and Zhang, Pingyue and Li, Linjie and Yang, Zhengyuan and Jin, Xing and Yu, Kefan and Nguyen, Minh Nhat and Liu, Licheng and others},
  journal={arXiv preprint arXiv:2504.20073},
  year={2025}
}

@article{shumailov2024ai,
  title={AI models collapse when trained on recursively generated data},
  author={Shumailov, Ilia and Shumaylov, Zakhar and Zhao, Yiren and Papernot, Nicolas and Anderson, Ross and Gal, Yarin},
  journal={Nature},
  volume={631},
  number={8022},
  pages={755--759},
  year={2024},
  publisher={Nature Publishing Group UK London}
}

@misc{openaio3,
  author = {OpenAI},
  year = {2025},
  url = {https://openai.com/index/introducing-o3-and-o4-mini/},
  title = {Introducing OpenAI o3 and o4-mini}
}

@misc{openaigpt5,
  author = {OpenAI},
  year = {2025},
  url = {https://openai.com/index/introducing-gpt-5/},
  title = {Introducing GPT-5}
}

@misc{qwen3,
  author = {Qwen Team},
  year = {2025},
  url = {https://qwen.ai/blog?id=1e3fa5c2d4662af2855586055ad037ed9e555125},
  title = {Qwen3: Think Deeper, Act Faster}
}

@misc{gemeni2.5pro,
  author = {Google Deepmind},
  year = {2025},
  url = {https://deepmind.google/models/gemini/pro/},
  title = {Gemini 2.5 Pro}
}

@article{yadav2023ties,
  title={Ties-merging: Resolving interference when merging models},
  author={Yadav, Prateek and Tam, Derek and Choshen, Leshem and Raffel, Colin A and Bansal, Mohit},
  journal={Advances in Neural Information Processing Systems},
  volume={36},
  pages={7093--7115},
  year={2023}
}

@article{bai2025qwen2.5,
  title={Qwen2.5-VL Technical Report},
  author={Bai, Shuai and Chen, Keqin and Liu, Xuejing and Wang, Jialin and Ge, Wenbin and Song, Sibo and Dang, Kai and Wang, Peng and Wang, Shijie and Tang, Jun and others},
  journal={arXiv preprint arXiv:2502.13923},
  year={2025}
}

@inproceedings{ilharcoediting,
  title={Editing models with task arithmetic},
  author={Ilharco, Gabriel and Ribeiro, Marco Tulio and Wortsman, Mitchell and Schmidt, Ludwig and Hajishirzi, Hannaneh and Farhadi, Ali},
  year={2023},
  booktitle={The Eleventh International Conference on Learning Representations}
}

@inproceedings{yangadamerging,
  title={AdaMerging: Adaptive Model Merging for Multi-Task Learning},
  author={Yang, Enneng and Wang, Zhenyi and Shen, Li and Liu, Shiwei and Guo, Guibing and Wang, Xingwei and Tao, Dacheng},
  year={2024},
  booktitle={The Twelfth International Conference on Learning Representations}
}

@inproceedings{yu2024language,
  title={Language models are super mario: Absorbing abilities from homologous models as a free lunch},
  author={Yu, Le and Yu, Bowen and Yu, Haiyang and Huang, Fei and Li, Yongbin},
  booktitle={Forty-first International Conference on Machine Learning},
  year={2024}
}

@inproceedings{wortsman2022model,
  title={Model soups: averaging weights of multiple fine-tuned models improves accuracy without increasing inference time},
  author={Wortsman, Mitchell and Ilharco, Gabriel and Gadre, Samir Ya and Roelofs, Rebecca and Gontijo-Lopes, Raphael and Morcos, Ari S and Namkoong, Hongseok and Farhadi, Ali and Carmon, Yair and Kornblith, Simon and others},
  booktitle={International conference on machine learning},
  pages={23965--23998},
  year={2022},
  organization={PMLR}
}

@article{yang2024model,
  title={Model Merging in LLMs, MLLMs, and Beyond: Methods, Theories, Applications and Opportunities},
  author={Yang, Enneng and Shen, Li and Guo, Guibing and Wang, Xingwei and Cao, Xiaochun and Zhang, Jie and Tao, Dacheng},
  journal={CoRR},
  year={2024}
}

@inproceedings{huang2017snapshot,
  title={Snapshot Ensembles: Train 1, Get M for Free},
  author={Huang, Gao and Li, Yixuan and Pleiss, Geoff and Liu, Zhuang and Hopcroft, John E and Weinberger, Kilian Q},
  booktitle={International Conference on Learning Representations},
  year={2017}
}

@article{li2025temporal,
  title={Temporal Sampling for Forgotten Reasoning in LLMs},
  author={Li, Yuetai and Xu, Zhangchen and Jiang, Fengqing and Ramasubramanian, Bhaskar and Niu, Luyao and Lin, Bill Yuchen and Yue, Xiang and Poovendran, Radha},
  journal={arXiv preprint arXiv:2505.20196},
  year={2025}
}

@article{li2025model,
  title={Model Merging in Pre-training of Large Language Models},
  author={Li, Yunshui and Ma, Yiyuan and Yan, Shen and Zhang, Chaoyi and Liu, Jing and Lu, Jianqiao and Xu, Ziwen and Chen, Mengzhao and Wang, Minrui and Zhan, Shiyi and others},
  journal={arXiv preprint arXiv:2505.12082},
  year={2025}
}

@inproceedings{sanyalearly,
  title={Early Weight Averaging meets High Learning Rates for LLM Pre-training},
  author={Sanyal, Sunny and Neerkaje, Atula Tejaswi and Kaddour, Jean and Kumar, Abhishek and others},
  booktitle={First Conference on Language Modeling},
  year={2024}
}

@inproceedings{guminillm,
  title={MiniLLM: Knowledge Distillation of Large Language Models},
  author={Gu, Yuxian and Dong, Li and Wei, Furu and Huang, Minlie},
  booktitle={The Twelfth International Conference on Learning Representations},
  yesr={2024}
}

@inproceedings{wu2025rethinking,
  title={Rethinking kullback-leibler divergence in knowledge distillation for large language models},
  author={Wu, Taiqiang and Tao, Chaofan and Wang, Jiahao and Yang, Runming and Zhao, Zhe and Wong, Ngai},
  booktitle={Proceedings of the 31st International Conference on Computational Linguistics},
  pages={5737--5755},
  year={2025}
}

@article{yu2025dapo,
  title={DAPO: An Open-Source LLM Reinforcement Learning System at Scale},
  author={Yu, Qiying and Zhang, Zheng and Zhu, Ruofei and Yuan, Yufeng and Zuo, Xiaochen and Yue, Yu and Fan, Tiantian and Liu, Gaohong and Liu, Lingjun and Liu, Xin and others},
  journal={CoRR},
  year={2025}
}

@article{fu2025areal,
  title={AReaL: A Large-Scale Asynchronous Reinforcement Learning System for Language Reasoning},
  author={Fu, Wei and Gao, Jiaxuan and Shen, Xujie and Zhu, Chen and Mei, Zhiyu and He, Chuyi and Xu, Shusheng and Wei, Guo and Mei, Jun and Wang, Jiashu and others},
  journal={arXiv preprint arXiv:2505.24298},
  year={2025}
}

@article{li2025efficient,
  title={Efficient multi-turn rl for gui agents via decoupled training and adaptive data curation},
  author={Li, Pengxiang and Hu, Zechen and Shang, Zirui and Wu, Jingrong and Liu, Yang and Liu, Hui and Gao, Zhi and Shi, Chenrui and Zhang, Bofei and Zhang, Zihao and others},
  journal={arXiv preprint arXiv:2509.23866},
  year={2025}
}

@article{zhang2025agentrl,
  title={AgentRL: Scaling Agentic Reinforcement Learning with a Multi-Turn, Multi-Task Framework},
  author={Zhang, Hanchen and Liu, Xiao and Lv, Bowen and Sun, Xueqiao and Jing, Bohao and Iong, Iat Long and Hou, Zhenyu and Qi, Zehan and Lai, Hanyu and Xu, Yifan and others},
  journal={arXiv preprint arXiv:2510.04206},
  year={2025}
}

@article{wang2025vagen,
  title={VAGEN: Reinforcing world model reasoning for multi-turn vlm agents},
  author={Wang, Kangrui and Zhang, Pingyue and Wang, Zihan and Gao, Yaning and Li, Linjie and Wang, Qineng and Chen, Hanyang and Wan, Chi and Lu, Yiping and Yang, Zhengyuan and others},
  journal={arXiv preprint arXiv:2510.16907},
  year={2025}
}

@article{feng2025group,
  title={Group-in-group policy optimization for llm agent training},
  author={Feng, Lang and Xue, Zhenghai and Liu, Tingcong and An, Bo},
  journal={arXiv preprint arXiv:2505.10978},
  year={2025}
}

@article{matena2022merging,
  title={Merging models with fisher-weighted averaging},
  author={Matena, Michael S and Raffel, Colin A},
  journal={Advances in Neural Information Processing Systems},
  volume={35},
  pages={17703--17716},
  year={2022}
}

@misc{schulmankldiv,
  author = {John Schulman},
  year = {2020},
  url = {http://joschu.net/blog/kl-approx.html},
  title = {Approximating KL Divergence}
}

@misc{jangrevkl,
  author = {Eric Jang},
  year = {2016},
  url = {https://blog.evjang.com/2016/08/variational-bayes.html},
  title = {A Beginner's Guide to Variational Methods: Mean-Field Approximation}
}

@article{cui2025entropy,
  title={The entropy mechanism of reinforcement learning for reasoning language models},
  author={Cui, Ganqu and Zhang, Yuchen and Chen, Jiacheng and Yuan, Lifan and Wang, Zhi and Zuo, Yuxin and Li, Haozhan and Fan, Yuchen and Chen, Huayu and Chen, Weize and others},
  journal={arXiv preprint arXiv:2505.22617},
  year={2025}
}

@article{xie2024osworld,
  title={Osworld: Benchmarking multimodal agents for open-ended tasks in real computer environments},
  author={Xie, Tianbao and Zhang, Danyang and Chen, Jixuan and Li, Xiaochuan and Zhao, Siheng and Cao, Ruisheng and Hua, Toh J and Cheng, Zhoujun and Shin, Dongchan and Lei, Fangyu and others},
  journal={Advances in Neural Information Processing Systems},
  volume={37},
  pages={52040--52094},
  year={2024}
}

@article{rawles2024androidworld,
  title={Androidworld: A dynamic benchmarking environment for autonomous agents},
  author={Rawles, Christopher and Clinckemaillie, Sarah and Chang, Yifan and Waltz, Jonathan and Lau, Gabrielle and Fair, Marybeth and Li, Alice and Bishop, William and Li, Wei and Campbell-Ajala, Folawiyo and others},
  journal={arXiv preprint arXiv:2405.14573},
  year={2024}
}

@article{rawles2023androidinthewild,
  title={Androidinthewild: A large-scale dataset for android device control},
  author={Rawles, Christopher and Li, Alice and Rodriguez, Daniel and Riva, Oriana and Lillicrap, Timothy},
  journal={Advances in Neural Information Processing Systems},
  volume={36},
  pages={59708--59728},
  year={2023}
}

@article{yue2025does,
  title={Does reinforcement learning really incentivize reasoning capacity in llms beyond the base model?},
  author={Yue, Yang and Chen, Zhiqi and Lu, Rui and Zhao, Andrew and Wang, Zhaokai and Song, Shiji and Huang, Gao},
  journal={arXiv preprint arXiv:2504.13837},
  year={2025}
}

@article{yuan2023scaling,
  title={Scaling relationship on learning mathematical reasoning with large language models},
  author={Yuan, Zheng and Yuan, Hongyi and Li, Chengpeng and Dong, Guanting and Lu, Keming and Tan, Chuanqi and Zhou, Chang and Zhou, Jingren},
  journal={arXiv preprint arXiv:2308.01825},
  year={2023}
}

@article{Qwen3-VL,
      title={Qwen3-VL Technical Report}, 
      author={Shuai Bai and Yuxuan Cai and Ruizhe Chen and Keqin Chen and Xionghui Chen and Zesen Cheng and Lianghao Deng and Wei Ding and Chang Gao and Chunjiang Ge and others},
	  journal={arXiv preprint arXiv:2511.21631},
      year={2025}
}
